\documentclass[]{new_tlp}
\makeatletter
\let\O@argtabularcr\@argtabularcr
\def\O@xtabularcr{\@ifnextchar[\O@argtabularcr{\ifnum 0=`{\fi}\cr}}
\let\O@tabacol\@tabacol
\let\O@tabclassiv\@tabclassiv
\let\O@tabclassz\@tabclassz
\let\O@tabarray\@tabarray
\def\author@tabular{\authorsize\def\@halignto{}\@authortable}
\let\endauthor@tabular=\endtabular
\def\author@tabcrone{{\ifnum0=`}\fi\O@xtabularcr\affilsize\itshape
 \let\\=\author@tabcrtwo\ignorespaces}
\def\author@tabcrtwo{{\ifnum0=`}\fi\O@xtabularcr[-3\p@]\affilsize\itshape
 \let\\=\author@tabcrtwo\ignorespaces}
\def\@authortable{\leavevmode \hbox \bgroup $\let\@acol\O@tabacol
 \let\@classz\O@tabclassz \let\@classiv\O@tabclassiv
 \let\\=\author@tabcrone \ignorespaces \O@tabarray}
\makeatother

\let\proof\relax

\usepackage[utf8]{inputenc}
\usepackage[english]{babel}

\usepackage{natbib}
\usepackage{tabularx,lipsum,environ,amsmath,amsthm,amssymb,float,forest,nicefrac, multirow, physics, color, courier, lineno, hyperref,comment,soul, xcolor,comment, xspace, colortbl}
\usepackage{soul}
\usepackage{listings}
\usepackage{graphicx}
\usepackage{hhline}
\usepackage{letltxmacro, changepage}
\usepackage{array}
\usepackage[ruled,vlined]{algorithm2e}
\newcolumntype{L}{>{\centering\arraybackslash}m{1.5cm}}
\newcolumntype{M}{>{\centering\arraybackslash}m{2cm}}
\newcolumntype{N}{>{\centering\arraybackslash}m{3cm}}
\newcolumntype{O}{>{\centering\arraybackslash}m{3.05cm}}
\newcolumntype{P}{>{\centering\arraybackslash}m{2.8cm}}

\LetLtxMacro\oldttfamily\ttfamily
\DeclareRobustCommand{\ttfamily}{\oldttfamily\csname ttsize\endcsname}
\newcommand{\setttsize}[1]{\def\ttsize{#1}}%

\newcommand{\dreaml}{\textit{DiceML}\xspace}

\setttsize{\small}
\newcounter{question}[section]
\newenvironment{question}[1][]{\refstepcounter{question}\par\medskip
   \textbf{Question~\thequestion. #1} \rmfamily}{\medskip}
   
\theoremstyle{definition}
\newtheorem{example}{Example}[section]

\theoremstyle{definition}
\newtheorem{definition}{Definition}[section]

\title[Theory and Practice of Logic Programming]
        {Learning Distributional Programs for Relational Autocompletion}

\author[Nitesh Kumar, Ond\v rej Ku\v zelka and Luc De Raedt]
     {Nitesh Kumar\\
     Department of Computer Science, KU Leuven, Belgium\\
     \email{nitesh.kumar@kuleuven.be}\\
     \and Ond\v rej Ku\v zelka\\
     Department of Computer Science, Czech Technical University in Prague, Czechia\\
     \email{ondrej.kuzelka@fel.cvut.cz}\\
     \and Luc De Raedt\\
     Department of Computer Science, KU Leuven, Belgium\\
     \email{luc.deraedt@kuleuven.be}}

\jdate{March 2003}
\pubyear{2003}
\pagerange{\pageref{firstpage}--\pageref{lastpage}}
\doi{S1471068401001193}

\begin{document}

\label{firstpage}
\maketitle
\begin{abstract}
    Relational autocompletion is the problem of automatically filling out some missing values in multi-relational data.
    We tackle this problem within the probabilistic logic programming framework of {\em Distributional Clauses} (DC), which supports both discrete and continuous probability distributions.  Within this framework, we introduce \dreaml \ -- an approach to learn both the structure and the parameters of DC programs from relational data (with possibly missing data).
    %, and that may also have missing information.  
    To realize this,  \dreaml integrates statistical modeling and distributional clauses with rule learning. %in a way that exploits the full expressiveness of DC and the capability of statistical models in learning intricate patterns. \dreaml learns both the structure and the parameters of such a distributional program, specifying a probability distribution over the entire database. 
    The distinguishing features of \dreaml are that it 1) tackles autocompletion in relational data,
    %\nit{autocompletion in multiple related tables},
    %relational autocompletion, 
    2) learns distributional clauses extended with statistical models, 3)  deals with both discrete and continuous distributions, 4) can exploit background knowledge, and 5) uses an expectation-maximization based algorithm to cope with missing data. 
    %In contrast with standard approaches to autocompletion, \dreaml is designed to deal with multiple related tables. In addition, it inherently handles discrete and continuous values and can make use of additional background knowledge, if available. %The problem becomes even more challenging if the database also has missing fields, and to the best of our knowledge has never been attempted before. 
    %To handle the missing fields in the database, \dreaml uses the stochastic expectation-maximization algorithm to learn the program. 
    The empirical results show the promise of the approach, even when there is missing data.

\end{abstract}


\section{Introduction}\label{intro}
Spreadsheets are arguably the most accessible tool for data analysis and millions of users use them. Generally, real-world data is not gathered in a single table but in multiple tables that are related to each other. Real-world data is often noisy and may have missing values. End users, however, do not have access to the state-of-the-art techniques offered by Statistical Relational AI \citep[StarAI,][]{kersting2011statistical} to analyze such data. To tackle this issue, we study the problem of
%introduce the task of 
{\em relational autocompletion}, where the goal is to automatically fill out the entries specified by users in multiple related tables.
%spreadsheets having multiple tables. 
%Relational databases are widely used to store real-world data. They consist of multiple tables containing information about various types of entities as well as the associated foreign keys, which capture relationships among them. Such real-world databases are often noisy and may have missing values. The relational autocompletion problem is to fill out (some of) these fields automatically. 
This problem setting is simple, yet challenging and is viewed as an essential component of an automatic data scientist \citep{DeRaedtIDA18}. We tackle this problem by learning a probabilistic logic program that defines the joint probability distribution over attributes of all instances in the multiple related tables. This program can then be used to estimate the most likely values of the cells of interest.

%The relational autocompletion problem is to fill out these missing values automatically, and that is viewed as a central problem in automated data science \citep{DeRaedtIDA18}. The problem can be solved by automatically learning a probabilistic relational model specifying a joint probability distribution over the entire database. The probabilistic inference in the learned model can then be used to estimate the most likely missing values. 

%The problem can be solved using statistical relational learning \cite{raedt2016statistical}\cite{koller2007introduction}\cite{neville2007relational}, by first defining a probabilistic relational model that specifies a joint probability distribution over the entire relational database, and then learning that model. The probabilistic inference in the learned model can be used to estimate the most likely missing values.  

%Once a probabilistic relational model has been learned, probabilistic inference can be used to compute the most likely missing values. 
%They can be used to enable auto-completion feature that deals with providing suggestions to users as they type a query. The feature used in software such as Excel can enhance the experience of millions of users. Many end-users lack the knowledge of machine learning models and can not design these models themselves. So, we need algorithms that learn the structure of these models automatically from the database. 

Probabilistic logic programming \citep[PLP,][]{ngo1997answering,sato1997prism,vennekens2004logic,Raedt2007ProbLogAP,poole2008independent} and statistical relational learning \citep[SRL,][]{Jaeger1997RelationalBN,richardson2006markov,koller2007introduction,natarajan2008learning,neville2007relational,Kimmig12ashort} have introduced various formalisms that integrate relational logic with graphical models.  While many PLP and SRL techniques exist, only a few of them are hybrid, i.e.,  can deal with both discrete and continuous variables. One of these hybrid formalisms are the {\em Distributional Clauses} (DC) introduced by  \cite{gutmann2011magic}. Distributional clauses form a probabilistic logic programming language that extends the programming language Prolog with continuous as well as discrete probability distributions.  It is this language that we adopt in this paper. 

We first integrate statistical models in distributional clauses and use these to learn intricate patterns present in the data. This extended DC framework allows us to learn a DC program that specifies a probability distribution over attributes of
%the entire database
multiple tables.
Just like graphical models, this program can then be used for various types of inference. For instance, one can infer not only the output of statistical models based on their inputs but also the input when the output is observed.

In line with inductive logic programming \citep{muggleton1991inductive, lavrac1994inductive, quinlan1995induction}, we propose an approach, named \dreaml\footnotemark\ (\textit{Di}stributional \textit{C}lauses with Statistical \textit{M}odels \textit{L}earner), that learns such a DC program from 
%a database 
relational data
and background knowledge. \dreaml jointly learns the structure of distributional clauses, the parameters of its probability distributions and the parameters of the statistical models. The learned program can subsequently be used for autocompletion.
\footnotetext{The code is publicly available: \url{https://github.com/niteshroyal/DiceML}, publication date: 15/09/19}

We study the problem also in the presence of missing data.
The problem of learning the structure of hybrid relational models then becomes even more challenging and has, to the best of our knowledge, never been attempted before. To tackle this problem, \dreaml performs structure learning inside the stochastic EM procedure \citep{diebolt1995stochastic}.

\paragraph{Related Work} There are several works in SRL for learning probabilistic models for relational data, such as probabilistic relational models \citep[PRMs,][]{friedman1999learning}, relational Markov networks \citep[RMNs,][]{taskar2002discriminative}, and relational dependency networks \citep[RDNs,][]{neville2007relational}. PRMs extend Bayesian networks with concepts of objects, their properties, and relations between them. RDNs extend dependency networks, and RMNs extend Markov networks in the same relational setting. However, these models are generally restricted to discrete data. To address this shortcoming, several hybrid SRL formalisms were proposed such as continuous Bayesian logic programs \citep[CBLPs,][]{kersting20071}, hybrid Markov logic networks \citep[HMLNs,][]{wang2008hybrid}, 
%hybrid probabilistic logic programming \citep[HProbLog,][]{gutmann2010extending}, 
hybrid probabilistic relational models \citep[HPRMs,][]{narman2010hybrid}, and relational continuous models \citep[RCMs,][]{choi2010lifted}. The work on hybrid SRL has mainly been focused on developing theory to represent continuous variables within the various SRL formalisms and on adapting inference procedures for hybrid domains. 
%\nit{Some} frameworks provide support for learning the parameters of a handcrafted structure from data. 
However, little attention has been given to the design of algorithms for structure learning of hybrid SRL models. The same is true for works on hybrid probabilistic programming \citep[HProbLog,][]{gutmann2010extending}, \citep[DC,][]{gutmann2011magic, nitti2016probabilistic}, \citep[Extended-Prism,][]{islam2012inference},  \citep[Hybrid-cplint,][]{alberti2017cplint}, \citep{michels2016approximate}, \citep[BLOG,][]{wu2018discrete}, \citep{pedro2019a}. Closest to our work is the work on hybrid relational dependency networks \citep[HRDNs,][]{ravkic2015learning}, for which structure learning was also studied, but this learning algorithm
%approach 
assumes that the data is fully observed. There are also few approaches for structure learning in the presence of missing data such as \cite{kersting2005say, khot2012structure, khot2015gradient}. However, these approaches are restricted to discrete data. 
%Furthermore, inference in hybrid models in probabilistic programming is not straightforward and has received some attention recently \citep{wu2018discrete}; however, these models are not relational. Inference in existing hybrid models that extend probabilistic graphical models with relations, such as HRDNs, is not well-studied \citep{ravkic2015learning}. 
Furthermore, existing hybrid models that extend probabilistic graphical models with relations, such as HRDNs, are associated with local probability distributions such as conditional probability tables.
%Additionally, HRDNs are associated with local probability distributions, such as conditional probability tables. 
As a result, it is difficult to represent certain independencies such as context-specific independencies \citep[CSIs,][]{boutilier1996context}. On the contrary, DC can represent CSIs leading to interpretable DC programs. 
%Recently, a constraint-learning system TaCLe \cite{kolb2017learning} was introduced to tackle the problem of auto-completion, however, the system only learns hard constraints and can not deal with noisy data. 
%Two recent probabilistic modeling languages BayesDB \citep{mansinghka2015bayesdb} and Tabular \citep{gordon2014tabular}, are designed for dealing with relational data in the tabular form, automatically fill out missing values in the table. However, they require the model to be defined by the user and do not learn the structure of the model. 
%Furthermore, existing hybrid models that extend probabilistic graphical models with relations, such as HRDNs, 
%except for HMLNs and \nit{hybrid PLPs},
%HProbLog, 
%Moreover, CSIs present in the program are exploited by the DC inference mechanism for efficient inference \citep{nitti2016learning}. \oknote{I don't completely understand this remark about CSIs. It is possible that the readers will not understand it either.}

%\oknote{\cite{kersting2005say} - the differences are similar as with the general structural EM and stochastic EM (described in the section about stochastic EM)... maybe we want to say it here again? or move this reference there?}
%\luc{this was only for logical hmms? a much less expressive framewor}
%\oknote{I was looking for PRMs with continuous distributions and so far I only found: \cite{choi2010lifted}. It is only for pairwise potentials (it is lifted) - because of that it is less expressive than DC.}

Learning meaningful and interpretable symbolic representations from data in the form of rules has been studied in many forms by the inductive logic programming(ILP) community \citep{quinlan1990learning, muggleton1995inverse, blockeel1998top, srinivasan2001aleph}. The standard ILP setting requires the input to be deterministic and usually the rules as well. Although some rule learners \citep{neville2003learning, vens2006remauve} output the confidence of their predictions, the rules learned for different targets have not been used jointly for probabilistic inference. To alleviate these limitations, \cite{de2015inducing} proposed ProbFoil+ that can learn probabilistic rules from probabilistic data and background knowledge. In this approach, rules learned for different targets can jointly be used for inference. However, this approach does not deal with continuous random variables and missing data. A handful approaches can learn rules with continuous probability distributions, and the learned rules can also be jointly used for inference. One such approach was proposed by \cite{speichert2018learning} using piecewise polynomials to learn intricate patterns from data. This approach differs from our approach as we use statistical models to learn these patterns.  Moreover, it is restricted to fully observed deterministic input. Another approach for structure learning of dynamic distributional clauses, an extended DC framework that deals with time, has also been proposed by \cite{nitti2016learning}. However, this approach cannot learn distributional clauses from background knowledge, which itself can be a set of distributional clauses. Furthermore, it learns the dynamic distributional clauses from fully observed data and does not deal with missing values
%data  
in relational data as we do.  To the best of our knowledge, the present paper makes the first attempt to learn interpretable hybrid probabilistic logic programs from partially observed probabilistic data as well as background knowledge. 
DC programs have been successfully applied in robotics and perceptual anchoring using handcrafted programs or by learning parameters of simple programs with defined structure \citep{moldovan2018relational, persson2019semantic}. The technique we present in the present
paper has already been successfully applied for structure learning in the perceptual anchoring context \citep{zuidberg} and extends these other results.

%but not by learning the structure of the programs as we do.
%databases. 

Our approach also deals with missing values in relational data. %\textcolor{red}{TODO: I am reading through some database cleaning/repair papers... will add more very soon, now just references and notes for myself: \cite{ilyas2015trends} (a review), \cite{rekatsinas2017holoclean} (they do not have continuous RVs), \cite{yakout2013don}.} 
Thus, it is also related to the vast literature on database {\em cleaning} \citep{ilyas2015trends}. However, there are not many database-cleaning methods that can learn distributions of the data and use them to automatically fill in missing data (mostly due to the complexity of the problem and the scale of real-world relational databases), and those methods that can to some extent model probability distributions, e.g.\ \citep{yakout2013don,rekatsinas2017holoclean}, still cannot model complex probability distributions involving both discrete and continuous random variables. While the approach presented in this paper cannot scale to databases containing billions of tuples, it can model very complex probabilistic distributions.

A different approach for autocompletion in spreadsheets was proposed by \citet{kolb2020predictive}. In this approach, multiple related tables are joined in a pre-processing step in order to obtain a single table, and then constraints and Bayesian networks are learned. Thus this approach propositionalizes the data,
which implies that the joined table may contain redundant information, implying that the learned model will not be succinct. Learning succinct first-order probabilistic models, which we do, is required to truly deal with relational data. 

\paragraph{Contributions} We summarise our contributions in this paper as follows:
\begin{itemize}
    \item We integrate distributional clauses with statistical models and use the resulting framework to represent a hybrid probabilistic relational model.
    \item We introduce \dreaml, the approach for relational autocompletion that learns distributional clauses with statistical models from %relational databases
    relational data
    and background knowledge.
    \item We extend \dreaml to learn DC programs from 
    %relational databases
    relational data
    with missing values 
    %data
    using the stochastic EM algorithm. 
    \item We empirically evaluate \dreaml on synthetic as well as real-world 
    %databases
    data, which shows the promise of our approach.
    %relational autocompletion based on learning the DC program from the relational database with missing values.
    %Representation of a hybrid probabilistic relational model in the DC formalism as a joint model program (JMP).
    %\item A learning algorithm that automatically learns the program from the relational database containing missing data. The program can then be used for the auto-completion task. 
    %\item \dreaml, an approach to relational autocompletion based on learning the DC program from the relational database with missing values.
    %\item An approximate EM algorithm for the relational autocompletion problem.
    %\item An extensive empirical evaluation on real-world databases.
\end{itemize}

\paragraph{Organization}
The paper is organized as follows. We start by sketching the problem setting in Section \ref{section: problem setting}. Section \ref{section: plp} reviews logic programming concepts and distributional clauses. In Section \ref{section: extended DC}, we discuss the integration of distributional clauses with statistical models. In Section \ref{section: JMP}, we describe the specification of the DC program that we shall learn. 
%a joint model program (JMP) which represents the hybrid probabilistic relational model in the form of a DC program. 
Section \ref{section: learning} explains the learning algorithm, which is then evaluated in  Section \ref{section: experiments}.

\section{Problem Setting}\label{section: problem setting}

Let us introduce {\em relational autocompletion} using the simplified spreadsheet in Table \ref{tab:database}.
%\nit{Non-expert users often gather their data in spreadsheets, which require a prepossessing step before standard machine learning algorithms can use them. We assume that such spreadsheets are already transformed into the right format using the automatic wrangling approach presented in \cite{verbruggen2018automatically}. For instance, the example in Table \ref{tab:database} is already in this format.} 
%Consider the example 
%%database 
%\nit{spreadsheet}
%shown in Table \ref{tab:database}. 
It consists of {\em entity tables} and {\em associative tables}. Each entity table (e.g., client, loan, and account) contains information about instances of the same type. An associative table (e.g., hasAcc and hasLoan) encodes a relationship among entities. 
%In principle, any relational database can be converted to this form of the database. 
This toy example illustrates two important properties of real-world applications, namely i) the attributes of entities may be numeric or categorical, and ii) there may be missing values in entity tables.  These %at random
are denoted by ``$-$''.

In addition, certain knowledge
%information 
is available
%known 
beforehand, and inclusion of this {\em background knowledge}
%information 
might be useful for learning; for instance, if a client of a bank has an account, and the account is linked to a loan, then the client has the loan. Knowledge
%information 
may even be uncertain; for instance, we might already have a probabilistic model that specifies a probability distribution over the age of clients.

The problem that we tackle in this paper is to autocomplete specific cells selected by users, 
%the user's queries 
denoted by ``$?$''.
%in the 
%database
%spreadsheet. 
%That is, the user can put a ``$?$'' in a field, and the task of the system is to fill out the right values.  We will call such ``$?$'' queries. 
This problem will be solved by automatically learning a DC program from such %database
%\nit{spreadsheet}.
data and background knowledge.
This program can then be used to fill out those cells
%the queries 
with the most likely values. This setting can be viewed as a simple nontrivial setting for automating data science \citep{DeRaedtIDA18}. 
%missing values and question marks. 

%\luc{Maybe we should name the database as an incomplete or a query  database -- as it is not a standard one with the question marks}
\begin{table}
\par
\parbox{0.33\textwidth}{
\centering
\begin{tabular}{|c|c|c|}
\hline
\multicolumn{3}{|c|}{\textbf{client}}\\
\hline
\textit{\underline{cliId}}& \textit{age} & \textit{creditScore}\\
\hline \hline
ann & 33 & $-$\\ \hline
bob & 40 & 500\\ \hline 
carl & $-$ & 450\\ \hline
john & 55 & 700\\
\hline
\end{tabular}
}
\parbox{0.3\textwidth}{
\centering
\begin{tabular}{|c|c|}
\hline
\multicolumn{2}{|c|}{\textbf{hasAcc}}\\
\hline
\textit{\underline{cliId}} & \textit{\underline{accId}}\\
\hline \hline
ann & a\_11\\ \hline
bob & a\_11\\ \hline
ann & a\_20\\ \hline 
john & a\_10\\
\hline
\end{tabular}
}
\parbox{0.3\textwidth}{
\centering
\begin{tabular}{|c|c|}
\hline
\multicolumn{2}{|c|}{\textbf{hasLoan}}\\
\hline
\textit{\underline{accId}} & \textit{\underline{loanId}}\\
\hline \hline
a\_11 & l\_20\\ \hline
a\_10 & l\_20\\ \hline
a\_20 & l\_31\\ \hline
a\_20 & l\_41\\
\hline
\end{tabular}
}

\par\bigskip
\parbox{.45\textwidth}{
\centering
\begin{tabular}{|c|c|c|}
\hline
\multicolumn{3}{|c|}{\textbf{loan}}\\
\hline
\textit{\underline{loanId}} & \textit{loanAmt} & \textit{status}\\
\hline\hline
l\_20 & 20050 & appr \\ \hline
l\_21 & $-$ & pend\\ \hline
l\_31 & 25000 & decl\\ \hline
l\_41 & 10000 & $-$\\
\hline
\end{tabular}
}
\parbox{.45\textwidth}{
\centering
\begin{tabular}{|c|c|c|}
\hline
\multicolumn{3}{|c|}{\textbf{account}}\\
\hline
\textit{\underline{accId}} & \textit{savings} & \textit{freq}\\
\hline \hline
a\_10 & 3050 & high\\ \hline
a\_11 & $-$ & low\\ \hhline{-|-|-|}
a\_19 & 3010 & \cellcolor[HTML]{D7DBDD}$?$ \\ \hhline{-|-|-|}
a\_20 & \cellcolor[HTML]{D7DBDD}$?$ & \cellcolor[HTML]{D7DBDD}$?$\\
\hhline{-|-|-|}
\end{tabular}
}
\par\bigskip
\caption{An example of a spreadsheet consisting of entity tables (client, loan and account), and associative tables (hasLoan and hasAcc). Missing cells are denoted by ``$-$'' and the cells of interest
%user's query is
are denoted by ``$?$''.}
\label{tab:database}
\end{table}

\section{Probabilistic Logic Programming}\label{section: plp}
In this section, we first briefly review logic programming concepts and then introduce DC which extend logic programs with probability distributions.

\subsection{Logic Programming}

An {\em atom} $p(t_1, \dots, t_n)$ consists of a predicate $p/n$ of arity $n$ and terms $t_1, \dots, t_n$. A {\em term} is either a constant (written in lowercase), a variable (in uppercase), or a functor applied to a tuple of terms. For example, \texttt{hasLoan(a\_1,L)}, \texttt{hasLoan(a\_1,l\_1)} and \texttt{hasLoan(a\_1,func(L))} are atoms and \texttt{a\_1}, \texttt{L}, \texttt{l\_1} and \texttt{func(L)} are terms. A {\em literal} is an atom or its negation. Atoms which are negated are called {\em negative atoms} and atoms which are not negated are called {\em positive atoms}. A {\em clause} is a universally quantified disjunction of literals. A {\em definite clause} is a clause which contains exactly one positive atom and zero or more negative atoms. In logic programming, one usually writes definite clauses in the implication form $h \leftarrow b_1, ..., b_n$ (where we omit 
%writing 
the universal quantifiers for ease of writing). Here, the atom $h$ is called {\em head} of the clause; and the set of atoms $\{b_1, ..., b_n\}$ is called {\em body} of the clause. A clause with an empty body is called a {\em fact}.
A {\em logic program} consists of a set of definite clauses.
%The rule states that if all atoms in the body are true then the head atom must also be true. 

\begin{example}\label{ex:ex1}
The clause $c$ $\equiv$ \texttt{clientLoan(C,L) $\leftarrow$ hasAccount(C,A), hasLoan(A,L)} is a definite clause. Intuitively, it states that \texttt{L} is a loan of a client \texttt{C} if \texttt{C} has an account \texttt{A} and \texttt{A} is associated to the loan \texttt{L}. 
\end{example}

A term, atom or clause, is \textit{ground} if it does not contain any variable. A {\em substitution} $\theta = \{V_1/t_1, ..., V_m/t_m\}$ assigns  terms $t_i$ to variables $V_i$. Applying $\theta$ to a term, atom or clause $e$ yields the term, atom or clause $e\theta$, where all occurrences of $V_i$ in $e$ are replaced by the corresponding terms $t_i$. A substitution $\theta$ is a {\em grounding} for $c$ if $c\theta$ is ground, i.e., contains no variables (when there is no risk of confusion we drop ``for $c$'').

\begin{example}
Applying the substitution $\theta = \{\texttt{C}/\texttt{c\_1}\}$ to the clause $c$ from Example \ref{ex:ex1} yields $c\theta$ which is  \texttt{clientLoan(c\_1,L) $\leftarrow$ hasAccount(c\_1,A), hasLoan(A,L)}.  %For instance, a substitution $\theta = \{C/\texttt{c\_1}, A/\texttt{a\_1}, L/\texttt{l\_1}\}$ is the grounding substitution. 
\end{example}

A substitution $\theta$ {\em  unifies}  two atoms $l_1$ and $l_2$ if $l_1\theta = l_2\theta$. Such a substitution is called a unifier. Unification is not always possible. If there exists a unifier for two atoms $l_1$ and $l_2$, we call such atoms {\em unifiable} and we say that {\em $l_1$ and $l_2$ unify}.

\begin{example}
The substitution $\theta = \{ \texttt{C}/\texttt{c\_1}, \texttt{M}/\texttt{L} \}$ unifies the atoms \texttt{clientLoan(c\_1,L)} and \texttt{clientLoan(C,M)}.

%\nit{NK: Might be some problem in this sentence}
%In some cases does not exist, e.g.\ there is no unification of atoms $p(a,X)$ and $p(b,Y)$.
\end{example}

The {\em Herbrand base} of a logic program P, denoted HB(P), is the set of all ground atoms which can be constructed using the predicates, function symbols and constants from the program P. A {\em Herbrand interpretation} is an assignment of truth-values to all atoms in the Herbrand base. A Herbrand interpretation $I$ is a model of a clause $h \leftarrow \mathcal{Q}$, if and only if, for all grounding substitutions $\theta$ such that $\mathcal{Q}\theta \subseteq I$, it also holds that $h\theta \in I$. 
%\luc{delete A clause $c$ (logic program LP) {\em entails} another clause $c'$ (logic program LP$'$), denoted as $c \models c'$ (LP $\models$ LP$'$), if and only if, each model of $c$ (LP) is also a model of $c'$ (LP$'$). } 

The {\em least Herbrand model} of a logic program P, denoted LH(P),
is the intersection of all Herbrand models of the logic program P, i.e., it consists of all ground atoms $f \in \textrm{HB}(\textrm{P})$ that are logically entailed by the logic program P. %; whenever a fact $f$ is logically entailed by a program $f$, we write LP $\models f$. 
The least Herbrand model of a program P can be generated by repeatedly applying the so-called $T_\textrm{P}$ operator until fixpoint. Let $I$ be the set of all ground facts in the program P. Starting from the set $I$ of all ground facts contained in P, the $T_\textrm{P}$ operator is defined as follows:
\begin{equation}
    T_\textrm{P}(I) = \{h\theta \mid h \leftarrow \mathcal{Q} \in \textrm{P}, \mathcal{Q}\theta \subseteq I, \textrm{where } \theta \textrm{ is a grounding substitution for } h \leftarrow \mathcal{Q}\},
\end{equation}
That is, if the body of a rule is true in $I$ for a substitution $\theta$, the ground head $h\theta$ must be in $T_\textrm{P}(I)$. It is possible to derive all possible true ground atoms using the $T_\textrm{P}$ operator recursively, until a fixpoint is reached ($T_\textrm{P}(I) = I$), i.e., until no more ground atoms can be added to $I$. 

Given a logic program P, an {\em answer substitution} to a {\em query}  %$Q$ 
of the form  $?-q_1, \dots, q_m$, where the $q_i$ are literals, is a substitution $\theta$ such that $q_1\theta, \dots, q_m\theta$
%all $q_i\theta$ 
is entailed by P, i.e., belongs to  LH(P). %The answer substitution $\theta$ is often inferred using SLD-resolution. The {\em answer set} of $q$ is the set of all correct answers to $q$. Logic programming is especially suitable for representing relational data and performing inference over the data.

\subsection{Distributional Clauses} \label{section: distributional clauses}
%\nit{NK: Commented part in this section might be wrong. Luc and Ondrej, complete this section. I will look into this at the last.}
DC is a natural extension of logic programs for representing probability distributions introduced by \cite{gutmann2011magic}.

\begin{definition}
A distributional clause is a rule of the form $h \sim \mathcal{D} \leftarrow b_1, ..., b_n$, where $\sim$ is a binary predicate used in infix notation,  $h$ is a random variable term, and ${\cal D}$ a distributional term.
\end{definition}
A distributional clause specifies that for each grounding substitution $\theta$ of the clause, the random variable $h\theta$ is distributed as ${\cal D}\theta$ whenever all $b_i\theta$ hold. So $h$ and ${\cal D}$ are terms belonging to the Herbrand universe denoting random variables $r(t_1, ... , t_n)$ and distributions  $d(u_1, ...,u_k)$ respectively.  Unlike regular terms in the Herbrand universe, the random variable functors $r$ and distribution functors $d$ cannot be nested. 

To refer to the values of the random variables, we use the binary predicate $\cong$, which is used in infix notation for convenience. Here, $r \cong v$ is defined to be true if $v$ is the value of the random variable $r$. 

\begin{example}
Consider the following distribution clause
% ground atoms  $h  \sim \mathcal{D}$ define the random variable $h$ as being distributed according to $\mathcal{D}$. 

% Note that the term $\mathcal{D}$ can  be non-ground.
%clientLoan(C,L) := hasAccount(C,A), hasLoan(A,L).  not a DC
\begin{lstlisting}[frame=none]
creditScore(C) ~ gaussian(755.5,0.1):=clientLoan(C,L),status(L)~=appr.
\end{lstlisting}

Applying the grounding substitution $\theta = \{\texttt{C}/\texttt{c\_1},\texttt{L}/\texttt{l\_1}\}$  to the distributional clause
%\begin{lstlisting}[frame=none] creditScore(C) ~ gaussian(755.5,0.1) := clientLoan(C,L), status(L)~=appr. \end{lstlisting}
 results in  defining the random variable \texttt{creditScore(c\_1)} as being drawn from the distribution $\mathcal{D}\theta = \texttt{gaussian(755.5, 0.1)}$ whenever \texttt{clientLoan(c\_1,l\_1)} is true and the outcome of the random variable \texttt{status(l\_1)} takes the value \texttt{appr} (``approved''), i.e., \texttt{status(l\_1)} $\cong$ \texttt{appr}. %For the substitution $\theta$, the distributional clause thus defines the conditional probability distribution: $p(h\theta \mid b_1\theta, ..., b_n\theta) = \mathcal{D}\theta$. 
\end{example}

\noindent A  distributional clause without body is called a {\em probabilistic fact}, e.g.
\begin{lstlisting}[frame=none]
age(c_2) ~ gaussian(40,0.2).
\end{lstlisting}

%The idea is that such ground atoms $h  \sim \mathcal{D}$ define the random variable $h$ as being distributed according to $\mathcal{D}$. %Here, $h \sim val(v)$ means that $h$ has value $v$ with probability $1$.
%\noindent The ground substitution $\theta$ of the clause $(h \sim \mathcal{D} \leftarrow b_1, ..., b_n)\theta$ in the semantics of the logic program, defines a random variable $h\theta$ drawn from the distribution $\mathcal{D}\theta$, whenever all $b_i\theta$ are true.
%The random variable $h\theta$ can be  discrete  or continuous. 
%\noindent The term $\mathcal{D}$ can also be non-ground as in the next distribution clause. %i.e., values, probabilities, or distribution parameters can be related to conditions in the body as in the clause $1$ of example \ref{example: DC with model} and $\mathcal{D}$ can also be ground as in the clause $8$ of example \ref{example: exampleDC}.
%%creditScore(C) ~ gaussian(350,0.1) := clientLoan(C,L), status(L)~=decl.
%\end{lstlisting}
%The key idea underlying the distributional-clause framework is that ground atoms correspond to random variables.
%They % we explain more details about the semantics further in the paper. 
%Hence, any ground atom can also be thought of as a name or a handle of a random variable. 
%When there is no risk of confusion we will refer to ground atoms and the respective random variables interchangeably. 
%To access the values of the random variables, we use the binary predicate $\cong$, which is used in infix notation for convenience.
%Here, $r \cong v$ is defined to be true if $v$ is the value of the random variable $r$. 
It is also possible to define random variables that take only one value with probability 1, i.e., {\em deterministic facts}, e.g.,
\begin{lstlisting}[frame=none]
age(c_1) ~ val(55).
\end{lstlisting}

A distributional program $\mathbb{P}$ consists of a set of distributional clauses and a set of  definite clauses.

The semantics of a distributional clause program is given by a set of {\em possible worlds},
which can be generated using the $ST_\mathbb{P}$ operator, a stochastic version of the $T_\textrm{P}$ operator. \cite{gutmann2011magic} define the $ST_\mathbb{P}$ operator using the following generative process. The process starts with an initial world $I$ containing all ground facts from the program. Then for each distributional clause  $h \sim \mathcal{D} \leftarrow b_1, ..., b_n$ in the program, whenever the body {$b_1\theta, ..., b_n\theta$} is true in the set $I$ for the grounding substitution $\theta$, a value $v$ for the random variable $h\theta$ is sampled from the distribution $\mathcal{D}\theta$ and $h\theta \cong v$ is added to the world $I$. This is also performed for deterministic clauses, adding ground atoms to $I$ whenever the body is true. 
%\oknote{I think this is a bit confusing. In \cite{gutmann2011magic}, there is the function READTABLE that, as far as I understand it, ensures that for each random variable, only one value is sampled. I don't think that it is clear from the description here.} 
A function \textsc{ReadTable($\cdot$)} keeps track of already sampled values of random variables and ensures that for each random variable, only one value is sampled.  
This process is then recursively repeated until a fixpoint is reached ($ST_\mathbb{P}(I) = I$), i.e., until no more variables can be sampled and added to the world.  The resulting world is called a {\em possible world}, while the intermediate worlds are called {\em partial possible worlds}. 

\begin{example}\label{example: possibleWorld}
Suppose that we are given the following DC program $\mathbb{P}$:

\begin{lstlisting}
hasAccount(c_1, a_1).
hasLoan(a_1, l_1).
age(c_1) ~ val(55).
age(c_2) ~ gaussian(40, 0.2).
status(l_1) ~ discrete([0.7:appr, 0.3:decl]). 
clientLoan(C,L) := hasAccount(C,A), hasLoan(A,L).
creditScore(C) ~ gaussian(755.5,0.1) := clientLoan(C,L), status(L)~=appr.
creditScore(C) ~ gaussian(350,0.1) := clientLoan(C,L), status(L)~=decl.
\end{lstlisting}

% \noindent Then 
% \begin{lstlisting}[frame=none]
% I = { hasAccount(c\_1,a\_1),  hasLoan(a\_1,l\_1), age(c\_1)=55, age(c\_2)=40.2, status(l\_1)=appr, clientLoan(c\_1,l\_1), creditScore(c\_1)=755.0 }
% \end{lstlisting}
% %\texttt{\{} \texttt{hasAccount(c\_1,a\_1)},  \texttt{hasLoan(a\_1,l\_1)}, \texttt{age(c\_1)=55}, \texttt{age(c\_2)=40.2}, \texttt{status(l\_1)=appr}, \texttt{clientLoan(c\_1,l\_1)}, \texttt{creditScore(c\_1)=755.0} \textt{\}} 
% is a possible world of the DC program $\mathbb{P}$.

%The unique least Herbrand model LH($\mathbb{P}$) is \big[\texttt{hasAccount(c\_1,a\_1)},  \texttt{hasLoan(a\_1,l\_1)}, \texttt{age(c\_1)}, \texttt{age(c\_2)}, \texttt{status(l\_1)}, \texttt{clientLoan(c\_1,l\_1)}, \texttt{creditScore(c\_1)}\big].

\noindent Applying the $ST_\mathbb{P}$ operator, we can sample a possible world of the program $\mathbb{P}$ as follows: 

\begin{lstlisting}[frame=none]
{hasAccount(c_1,a_1), hasLoan(a_1,l_1), age(c_1)~=55} ->
{hasAccount(c_1,a_1), hasLoan(a_1,l_1), age(c_1)~=55, age(c_2)~=40.2} ->
{hasAccount(c_1,a_1), hasLoan(a_1,l_1), age(c_1)~=55, age(c_2)~=40.2, status(l_1)~=appr} ->
{hasAccount(c_1,a_1), hasLoan(a_1,l_1), age(c_1)~=55, age(c_2)~=40.2, status(l_1)~=appr, clientLoan(c_1,l_1)} ->
{hasAccount(c_1,a_1), hasLoan(a_1,l_1), age(c_1)~=55, age(c_2)~=40.2, status(l_1)~=appr, clientLoan(c_1,l_1), creditScore(c_1)~=755.0}
\end{lstlisting} %The intermediate worlds are called {\em partial possible worlds}.
\end{example}

A distributional program $\mathbb{P}$ is {\em valid}, as mentioned in  \cite{gutmann2011magic}, if it satisfies the following conditions.
%\textcolor{red}{Are you sure the following is precise enough?} 
First, for each random variable $h\theta$, $h\theta \sim \mathcal{D}\theta$ has to be unique in the least fixpoint, i.e., %\oknote{least fixpoint? did you mean just ``fixpoint''?} 
there is one distribution defined for each random variable. Second, the program $\mathbb{P}$ needs to be stratified, i.e., there exists a rank assignment $\prec$ over predicates of the program such that for each distributional clause $h \sim \mathcal{D} \leftarrow b_1, ..., b_n : $ $b_i \prec h$, and for each definite clause $h \leftarrow b_1, ..., b_n : b_i \preceq h$.  Third, all ground probabilistic facts are Lebesgue-measurable. Fourth, each atom in the least fixpoint can be derived from a finite number of probabilistic facts. 

The first requirement is actually enforcing mutual exclusiveness for different ground rules defining the same random variable $h$; i.e.,
it enforces that the condition parts of the two rules are mutually exclusive. This is similar to the conditions imposed in  PRISM \citep{sato2001parameter}.  To understand this problem,  reconsider Example \ref{example: possibleWorld}. Suppose we add a fact \texttt{hasLoan(a\_1,l\_2)} in the DC program. The client \texttt{c\_1} now has two loans, namely, \texttt{l\_1} and \texttt{l\_2}. Suppose in a possible world the status of loan \texttt{l\_1} and \texttt{l\_2} are \texttt{decl} (``declined'') and \texttt{appr} (``approved'') respectively. There are thus, two different Gaussian distributions defined for the client score of \texttt{c\_1} in the world. The presence of two distributions for a single random variable violates the first validity condition  of DC programs. Therefore this situation is not allowed.
%\luc{reproduce proposition 1 from Gutmann}
\cite{gutmann2011magic} show that: when a distributional program $\mathbb{P}$ satisfies the {\em validity conditions} then $\mathbb{P}$ specifies a proper probability measure over the set of fixpoints of the operator $ST_\mathbb{P}$. 
%\begin{proposition}\label{proposition: DC}
%\nit{NK: Not sure whether it makes sense to keep the proposition here. }
%Let $\mathbb{P}$ be a valid program. $\mathbb{P}$ defines a probability measure $P_\mathbb{P}$  over the set of fixpoints of the operator $ST_\mathbb{P}$. Hence, $\mathbb{P}$ also defines, for an arbitrary formula q over atoms in its Herbrand base, the probability that q is true.
%\end{proposition}
%\noindent This proposition states that one obtains a proper probability measure when the distributional program satisfies the {\em validity conditions}.
%\nit{To define a consistent probability distribution $p(x)$, a DC

Inference in DC is the process of computing probability of a query $q$ given evidence $e$. Sampling full possible worlds for inference is generally inefficient or may not even terminate as possible worlds can be infinitely large. Therefore, DC uses an efficient sampling algorithm based on backward reasoning and likelihood weighting to generate only those facts that are relevant to answer the given query. To estimate the probability, samples of partial possible worlds, i.e., the set of {\em relevant} facts, are generated. 
%Backward reasoning is used to prove the query and the evidence. 
A partial possible world is generated after a successful completion of a proof of the evidence and the query using backward reasoning. The proof procedure is repeated $N$ times to estimate the probability $p(q \mid e)$ that is given by,
\begin{equation} \label{equation: probability weight}
p(q \mid e) = \frac{\sum_{i=1}^{N}w_{q}^{(i)} w_{e}^{(i)}}{\sum_{i=1}^{N}w_{e}^{(i)}} 
\end{equation}
where $w_e^{(i)}$ is the likelihood of $e$ in an $i^{th}$ sample of a partial possible world, and $w_q^{(i)}$ is $1$ if the world entails $q$;
%$q$ is true in the world; 
otherwise, it is $0$.
%the likelihood weight of $q$ in $i^{th}$ proof
(see \cite{nitti2016probabilistic} for details).

\section{Advanced Constructs in the DC Framework}\label{section: extended DC}

In this section, we describe three advanced modeling constructs 
%that can be represented 
in the DC framework. We allow for {\em negation}, {\em aggregation functions} and {\em statistical models} in bodies of the distributional clauses. 
%First, we allow aggregates in bodies of the distributional clauses and, second, we allow {\em statistical models} to be present in the bodies of the distributional clauses as well.
%to allow for aggregates as well as the type of mathematical models (especially those based on linear model) that we will use in our learning algorithm.

\subsection{Negation}
Following \cite{nitti2016probabilistic}, we also allow for 
 {\em negated literals} in the  body of distributional clauses, where negation is interpreted as negation as failure. 
 For instance:
\begin{lstlisting}[frame=none]
creditScore(C)~gaussian(855.5,0.2):=clientLoan(C,L),\+status(L)~=appr.
\end{lstlisting}
Here, the negation will succeed if the status of the loan \texttt{L} is anything but \texttt{appr}. 
It is also possible to use negation to refer to undefined variables, e.g.
when the status is undefined, one could use:
\begin{lstlisting}[frame=none]
creditScore(C) ~ gaussian(755.5,0.1) := clientLoan(C,L), \+status(L)~=_.
\end{lstlisting}
the comparison involving {\em undefined} status will fail,  thus its negation will succeed.

\subsection{Aggregation}\label{sec:aggregation}

%\luc{this is not the problem, we just want to aggregate and we do not want to bring 
%combining rules into the picture ... as in BLPs and PRMs}
%\section{Expressing relations and mathematical models in DC}\label{section: extending DC}
%\subsection{Handling relations}\label{section: handling relations}
%Often in relational domains, an instance of an entity is related to a set of instances of other type of entity and properties of the instance may depend on properties of other instances in the set. 
%Aggregation functions are used to combine the properties of instances together.  Examples include the mode (...), sum (...), mean (...), minimum, maximum, and cardinality (...) functions.  DC implements them via second order aggregation predicates  ... of the form aggr(T, Q, R) ...

The example about mutual exclusiveness, as discussed in Section \ref{section: distributional clauses}, points to the difficulty
of using the status of multiple loans in the basic version of the distributional clauses. Therefore, we introduce aggregation functions into distributional clauses.

%The presence of relations 
%such as \texttt{clientLoan(C,L)} 
%in the body of distributional clauses can be problematic. To understand this problem, let us reconsider Example \ref{example: possibleWorld}. Suppose we add a fact \texttt{hasLoan(a\_1,l\_2)} in the DC program. The client \texttt{c\_1} now has two loans, namely, \texttt{l\_1} and \texttt{l\_2}. Suppose in a possible world the status of loan \texttt{l\_1} and \texttt{l\_2} are \texttt{decl} \nit{(``declined'')} and \texttt{appr} \nit{(``approved'')} respectively. There are thus, two different Gaussian distributions defined for the client score of \texttt{c\_1} in the world. The presence of two distributions for a single random variable violates the first validity condition\footnotemark\ of DC programs.

%\footnotetext{In the case a ground atom can be derived from more than one ground rule, ProbLog combines the contributions in terms of probability of the ground rules with a noisy-OR \citep{NoisyOr}. Since ProbLog deals with only Boolean random variables combining rules with noisy-OR is natural. However, the generalization of such a combining rule, along with its proper implementation, for hybrid domains is still an open problem. In such a case, the current implementation of DC will naively select the first distribution from the list of distributions.}

%To avoid such a problem, we allow
Aggregation functions  
%Aggregation functions are used to 
combine the properties of a set of instances of a specific type into a single property. Examples include the mode (most frequently occurring value);  mean value (if values are numerical),  maximum or minimum, cardinality, etc. 
%An aggregation function defines how certain properties of a set of instances of certain type are combined into a single property. 
They are implemented by second order {\em aggregation predicates} in the body of clauses. Aggregation predicates are analogous to the $findall$ predicate in Prolog. They are of the form $aggr(T,Q,R)$, where $aggr$ is an aggregation function (e.g. sum), $T$ is the target aggregation variable that occurs in the conjunctive goal query $Q$, and $R$ is the result of the aggregation. 
%The goal query $Q$ is conjunction of atoms $(r_1, \dots, r_k, a_1)$, where $r_i$ is a predicate of the form $p(V_1, \dots, V_n)$ with variables $V_1, \dots, V_n$ and $a_1$ is the binary predicate $\cong$ that unifies the property of an instance with target aggregation variable $T$.
%Essentially, an aggregation predicate collects all answer substitutions for the variable $T$ in a list, then it applies the aggregation function $aggr$ on this list and it returns $R$. %The predicate can be used to avoid problems with the validity condition \cite{gutmann2011magic} of distributional clauses that states that there must be a unique ground distribution $\mathcal{D}$ for each ground random variable $h$. 
%Many useful aggregation functions can be used:  mode (most frequently occurring value);  mean value (if values are numerical),  maximum or minimum,   cardinality etc. 
\begin{example}
Consider the following two clauses:
\begin{lstlisting}[frame=none]
creditScore(C) ~ gaussian(755.5,0.1) := mod(T, (clientLoan(C,L), status(L)~=T), X), X==appr.
creditScore(C) ~ gaussian(500.5,0.1) := \+ mod(T, (clientLoan(C,L), status(L)~=T), X).
\end{lstlisting}%creditScore(C) ~ gaussian(350,0.1) := mod(T, (clientLoan(C,L), status(L)~=T), decl).
%\luc{This is a bit strange --- I would expect that you return a variable R and then test whether it is equal to appr to save computations.}
%\luc{this last sentence is not correct. Eihter mod succeeds, and then X will have a value, or it does not, and then never reaches the last literal}
The aggregation predicate \texttt{mod} in the body of the first clause collects the status of all loans that a client has into a list and unifies the constant \texttt{appr} (``approved'') with the most frequently occurring value in the list. Thus, the first clause's body will be true if and only if the most frequently occurring value in this list is \texttt{appr} (i.e., the clause will fire for those clients whose most loans are approved). It may also happen that a client has no loan, or the client has loans but the statuses of these loans are not defined. In this case, this aggregate predicate will fail,   and the body of the second clause will be true. 
\end{example}

\subsection{Distributional Clauses with Statistical Models} 
Next we look at the way continuous random variables can be used in the body of a distributional clause for specifying the distributions in the head. One possibility described in \cite{gutmann2011magic} is to use standard comparison operators in the body of the distributional clauses, e.g., $\geq, \leq, >, < $, which can be used to compare values of random variables with constants or with values of other random variables. 
%We will not allow comparison operators on continuous random variables in distributional programs that we will be learning in this paper.

Another possibility which we describe in this section, is to use a {\em statistical model} that maps outcomes of the random variables in the body of a distributional clause to parameters of the distribution in the head.
%The model implements a mathematical function that maps continuous random variables in the body to parameters of the distribution in the head.
%\begin{definition} (DC with mathematical model) 
Formally, a distributional clause with a statistical model is a rule of the form $h \sim \mathcal{D}_\phi \leftarrow b_1, ..., b_n, \mathcal{M}_\psi$, where $\mathcal{M}_\psi$ is an atom implementing a function with parameters $\psi$ which relates the continuous variables in $\{b_1, ..., b_n\}$ with parameters $\phi$ in the distribution $\mathcal{D}_\phi$. We allow for the statistical model atoms defined in Table \ref{table: dc with statistical models}.
\begin{table}[!ht]
\centering
 \begin{tabular}{|L|N|M|O|P|} 
 \hline
 type of random variable ($X$) & statistical model atom ($\mathcal{M}_\psi$) in the body & function implemented by $\mathcal{M}_\psi$ & the head & probability distribution/density of $X$\\ [0.5ex] 
 \hline\hline
 continuous & $\begin{aligned}&linear([Y_1,\dots, Y_n], \\ & [W_1,\dots, W_{n+1}],M) \end{aligned}$ & $\begin{aligned} M = Z \end{aligned}$ & $\begin{aligned} X \sim gaussian(M,\sigma^2) \end{aligned}$ & $\begin{aligned} & \frac{1}{\sigma \sqrt{2\pi}}e^{-\frac{1}{2\sigma^2}(X - M)^2} \end{aligned}$ \\ \hline
 boolean & $\begin{aligned}& logistic([Y_1,\dots, Y_n], \\&[W_1,\dots, W_{n+1}],\\& [P_1, P_2])\end{aligned}$ & $\begin{aligned}& P_1 = \frac{1}{1+e^{-Z}} \\& P_2 = 1 - P_1\end{aligned}$ & $X \sim discrete([P_1:true, P_2:false])$ & $P_1^{I\left[X=true\right]} \times P_2^{I\left[X=false\right]}$ \\ \hline
 discrete & $\begin{aligned}& softmax([Y_1,\dots,Y_n], \\& [[W_{1_1},\dots,W_{{n+1}_1}], \dots\\& , [W_{1_d},\dots,W_{{n+1}_d}]], \\& [P_1,\dots,P_d])\end{aligned}$ & $\begin{aligned}&P_1 = \frac{e^{Z_1}}{N} \\ &\vdots \\ & P_d = \frac{e^{Z_d}}{N}\end{aligned}$ & $X \sim discrete([P_1:l_1,\dots,P_d:l_d])$ & $P_1^{I\left[X=l_1\right]} \times \dots \times
P_d^{I\left[X=l_d\right]}$
 \\ \hline
  \multicolumn{5}{|c|}{where $Z$ is $Y_1 . W_1 + \dots + Y_n . W_n + W_{n+1}$,} \\
  \multicolumn{5}{|c|}{ $Z_i$ is $Y_1 . W_{1_i} + \dots + Y_n . W_{n_i} + W_{{n+1}_i}$,} \\
  \multicolumn{5}{|c|}{$N$ is $\sum_{i=1}^{d} e^{Z_i}$,} \\
  \multicolumn{5}{|c|}{$d$ is the size of domain of $X$ ($dom(X)$) and $l_i \in dom(X)$,} \\
  \multicolumn{5}{|c|}{and $I\left[\cdot\right]$ is the ``indicator function'', so that $I\left[\textrm{a true statement}\right]=1$, and} \\ 
  \multicolumn{5}{|c|}{$I\left[\textrm{a false statement}\right]=0$}\\\hline
\end{tabular}
\caption{The table specifies the functions implemented by various statistical model atoms ($\mathcal{M}_{\psi}$) in distributional clauses. The functions together with the distribution ($\mathcal{D}_{\phi}$) in the head of the clauses specify the probability distribution/density of the random variable ($X$) defined by the head.}
\label{table: dc with statistical models}
\end{table}

\begin{example}\label{example: DC with model}
Consider the following distributional clauses, which state that the credit score of a client depends on the age of the client. The loan status, which can either be high or low, depends on the amount of the loan. The loan amount is, in turn, distributed according to a Gaussian distribution. 
%Clause 1 is the corresponding DC with a linear model. Additionally, loan status that can take value high or low depends on the amount of the loan. Clause 2 is the corresponding DC with a logistic model.
\begin{lstlisting}[frame=none]
creditScore(C) ~ gaussian(M,0.1) := age(C)~=Y, linear([Y],[10.1,200],M).
status(L) ~ discrete(P1:low,P2:high) := loan(L), loanAmt(L)~=Y, logistic([Y], [1.1,2.0],[P1,P2]).
loanAmt(L) ~ gaussian(25472.3,10.2) := loan(L).
\end{lstlisting}
Here, in the first clause, the linear model atom
%\footnote{\texttt{linear([Y],[10.1,200],M)} implements the linear function \texttt{M is 10.1 $\cdot$ Y + 200}. \nit{It together with the distribution in the head} signifies that a credit score $X$ \nit{of a client} is sampled from a probability distribution $\frac{1}{\sqrt{0.2\pi}}e^{-\frac{1}{0.2}(X - 10.1\cdot Y - 200)^2}$}\
with parameters $\psi = [10.1, 200]$ relates the continuous variable $Y$ and the mean $M$ of the Gaussian distribution in the head. %\oknote{consider writing explicitly what the distribution will be in mathematical form with the actual constants, not just in the logic programming form, it is also a bit hidden in the footnotes}
Likewise, in the second clause, the logistic model atom
%\footnote{\texttt{logistic([Y],[1.1,2.0],[P1,P2])} implements the logistic function  \texttt{P1 is $\frac{1}{1+e^{ -1.1 \cdot Y - 2.0}}$}, \texttt{P2 is 1-P1}. \nit{It together with the distribution in the head} signifies that a status $X$ \nit{of a loan} is sampled from a probability distribution $\left\{\frac{1}{1+e^{ -1.1 \cdot Y-2.0}}\right\}^{1\left[X=low\right]}\left\{1-\frac{1}{1+e^{ -1.1 \cdot Y-2.0}}\right\}^{1\left[X=high\right]}$, where $1\left[\cdot\right]$ is the ``indicator function'', so that $1\left[\textrm{a true statement}\right]=1$, and $1\left[\textrm{a false statement}\right]=0$}
with parameter $\psi = [1.1, 2.0]$ relates $Y$ to the parameters $\phi = [P1,P2]$ of the discrete distribution in the head. %The model atom is essentially a built-in Prolog predicate that is supported by the DC inference engine. 
\end{example}

It is worth spending a moment studying the form of distributional clauses with statistical models as discussed above. Statistical models such as linear and logistic regression are fully integrated with the probabilistic logic framework in a way that exploits the full expressiveness of logic programming and the strengths of these models in learning intricate patterns. Moreover, we will see in Section \ref{section: learning} that these models can easily be learned along with the structure of the program. In this fully integrated framework, we not only infer in the forward direction, i.e., the output based on the input of these models but we can also infer in the backward direction, i.e., the input if we observe the output. For instance, in the above example, if we observe the status of the loan, then we can infer the loan amount, which is the input of the logistic model. Now, we can specify a complex probability distribution over continuous and/or discrete random variables using a distributional program having multiple clauses with statistical models. 

%\luc{why do not we replace the logistic and linear prediate in the clause 1 ad 2 by the formula ... easier to read and then you add a footnote -- for ease of  reading we represent the equations explicitly rather than defining them trhough extra predicates} \nit{NK: Prolog is slow for arithmetic operations. These predicates will be in C language. Complex functions can be implemented in predicates.}

% \luc{the rest of this section can be deleted} 
% In a clause of the form $h \sim \mathcal{D}_\phi \leftarrow b_1, ..., b_n, \mathcal{H}_\psi$, the distribution $\mathcal{D}_\phi$ in combination with the model atom $\mathcal{H}_\psi$ specifies a well-defined  probabilistic model $\mathcal{M}_\psi$ with the parameter $\psi$. Fundamentally, this means that the grounding substitution $\theta$ of the clause defines the random variable $h\theta$ drawn from the distribution defined by the {\em ground probabilistic model} $\mathcal{M}_\psi \theta$. Notice that the linear model in combination with Gaussian distribution in clause 1 of example \ref{example: DC with model} specifies a probabilistic linear regression model. Likewise, the logistic model in combination with discrete distribution in clause 2 specifies a probabilistic logistic regression model. We will see in the next section how DC with the mathematical model is useful in defining a joint model for a relational database.

\section{Joint Model Program for Multi-Relational Tables}\label{section: JMP}

We will now use the DC formalism to define a probability distribution over all attributes of multiple related tables.
%the entire relational database. 
The next subsections describe: (i) how to map tables 
%the relational database 
onto the set of distributional clauses, and (ii) the type of probabilistic relational model that we shall learn. %The next section then introduces our learning algorithm.

% \luc{DELETE .... 
% In this section, we introduce a joint relational model that defines a joint probability distribution over an entire database. The model extends PRM \cite{getoor2007probabilistic} like aggregation functions with mathematical models for handling continuous data. The model is represented in the form of a DC program, which we call a {\em joint model program} (JMP). Before discussing JMP, we discuss the representation of a database in DC that helps to formalize JMP.  
% I do not see whay we need to call this JMP -- it is just DC ... 
% }

%\subsection{Representation of a database}\label{section: representation}

\subsection{Modeling the Input Tables
%Spreadsheet
%Database 
(Sets $\mathcal{A}_{\mathcal{DB}}$ and $\mathcal{R}_{\mathcal{DB}}$)}\label{sec:modeling database}
In this paper, we use relational data
%spreadsheets
%relational databases 
consisting of multiple entity tables and multiple associative tables. The entity tables are assumed to contain no foreign keys whereas the associative tables are assumed to contain only foreign keys which represent relations among entities. Although this is not a standard form, any relational data
%database 
can be transformed into this canonical form, without loss of generality. For instance, data
%the spreadsheet
%database 
in Table \ref{tab:database} is already in this form. %The set of all instances in an entity table $e$ constitutes the {\em population} $pop(e) = \{t_1,\dots,t_n \}$. -> It is probably not used

Next, we transform the given relational data
%spreadsheet
%database 
$\mathcal{DB}$ to a set $\mathcal{A}_\mathcal{DB} \cup \mathcal{R}_\mathcal{DB}$ of facts that will be used as the training data.
%probabilistic and deterministic facts. 
Here, $\mathcal{A}_\mathcal{DB}$ contains information about the values of attributes, 
%represented using \nit{deterministic facts,} 
%probabilistic facts, 
and $\mathcal{R}_\mathcal{DB}$ consists of information about the relational structure of data
%the spreadsheet
%database 
(which entities exist and the relations among them).
%represented using \nit{usual} facts.
%various kinds of predicates and distributional clauses.
%\luc{

In particular, given
%spreadsheet
%a database 
$\mathcal{DB}$, we transform it as follows:

%\noindent {\bf \okedit{Construction of the sets $\mathcal{A}_\mathcal{DB}$ and $\mathcal{R}_\mathcal{DB}$}:}
\begin{itemize}
    \item For every instance $t$ in an entity table $e$, we add the %deterministic 
    fact $e(t)$ to $\mathcal{R}_\mathcal{DB}$. For example, from the client table, we add \texttt{client(ann)} for the instance \texttt{ann}.
    \item For each associative table $r$, we add 
    %deterministic 
    facts $r({t_1}, {t_2})$ to $\mathcal{R}_\mathcal{DB}$ for all tuples $({t_1}, {t_2})$ contained in the table $r$. For example, \texttt{hasAcc(ann,a\_11)}. 
    \item For each instance $t$ with an attribute $a$ of value $v$, we add a deterministic
    %probabilistic 
    fact $a(t) \sim val(v)$ to $\mathcal{A}_\mathcal{DB}$. For example, \texttt{age(ann) $\sim$ val(33)}.
    %For every attribute $a$ in the entity table, we introduce an {\em attribute predicate} of the form $a(T)$, e.g., \texttt{age(C)}.  %Here \texttt{val(33)} is a degenerate distribution in which $33$ has probability $1$ (this is a natural way to represent deterministic values in DC). If the value of the attribute is missing for the entity $t_i$ from the database then no distributional fact is added for it but, instead, we add just the corresponding logical atom, e.g. \texttt{age(c\_3)}. %(intuitively this means that the random variable \texttt{age(c\_3)} exists but its distribution is not known).\luc{What does that mean?}
\end{itemize}
We call $e/1$ the entity relation, $a/1$ an attribute, and  $r/2$ a link relation.

This representation of $\mathcal{DB}$
%spreadsheets
%databases 
ensures that the {\em existence}\footnotemark\ of the individual entity is not a random variable. Likewise, the relations among entities are also not random variables. On the other hand, attributes of instances 
%the values of attributes 
are random variables. For instance, in the preceding example \texttt{age(ann)} is a random variable. This is exactly what we need for the relational autocompletion setting that we study in this paper in which we are only interested in predicting missing values of attributes but not in predicting missing relations or missing entities.

The background knowledge ${\cal BK}$, if present, is written in the form of a set of distributional clauses and is used in training.

\footnotetext{Note that DC can represent uncertain existence and uncertain relations, as discussed in \citep{nitti2017planning}. However, the problem of learning existence is not well defined, and for learning a relation, we need both true and false examples of the relation. In the real world, we do not observe false examples, so learning relations is considered as a PU learning problem \citep{bekker2020}.}

\subsection{Modeling the Probability Distribution}
Next, we describe the form of DC programs, {\em joint model programs} (JMPs), that we will learn for the relational autocompletion problem.

A JMP  learned for a relational database ${\cal DB}$ consists of
\begin{enumerate}
    \item the facts in the transformed ${\cal R}_{\cal DB}$; 
    % \luc{should ${\cal A}_{\cal DB}$ not be inlcuded here too?}
    % \item possibly a background theory ${\cal BK}$ in the form of a set of distributional clauses;
    \item a set of learned distributional clauses ${\cal H}$ that together define all the attributes in the database.
\end{enumerate}
Furthermore, the learned clauses do not target relations and do not contain comparison operators, even though continuous random variables may affect other random variables via distributional clauses using statistical models. Observe that $\mathcal{A}_{\mathcal{DB}}$ does not belong to JMPs since it is used to train them.

\begin{example}\label{example: examplePRT1}
A JMP shown below specifies a distribution over all attributes of each instance in Table \ref{tab:database}. 
\begin{lstlisting}
client(ann).  client(john).  ...
hasAcc(ann,a_11).  hasAcc(ann,a_20).  ...
freq(A) ~ discrete([0.2:low,0.8:high]) := account(A).
savings(A) ~ gaussian(2002,10.2) := account(A), freq(A)~=X, X==low. 
savings(A) ~ gaussian(3030,11.3) := account(A), freq(A)~=X, X==high.
age(C) ~ gaussian(Mean,3) := client(C), avg(X,(hasAcc(C,A), savings(A)~=X), Y), creditScore(C)~=Z, linear([Y,Z],[30,0.2,-0.4],Mean). 
loanAmt(L) ~ gaussian(Mean,10) := loan(L), avg(X,(hasLoan(A,L), savings(A)~=X),Y), linear([Y],[100.1, 10],Mean).
loanAmt(L) ~ gaussian(25472.3,10.2) := loan(L), \+avg(X,(hasLoan(A,L),savings(A)~=X),Y).
status(L) ~ discrete([P1:appr, P2:pend, P3:decl]) := loan(L), avg(X, (hasLoan(A,L),hasAcc(C,A),creditScore(C)~=X),Y), loanAmt(L)~=Z, softmax([Y,Z],[[0.1,-0.3,-2.4],[0.3,0.4,0.2],[0.8,1.9,-2.9]],[P1,P2,P3]).
creditScore(C) ~ gaussian(300,10.1) := client(C), mod(X,(hasAcc(C,A), freq(A)~=X),Z), Z==low.
creditScore(C) ~ gaussian(Mean,15.3) := client(C), mod(X,(hasAcc(C,A), freq(A)~=X),Z), Z==high, max(X,(hasAcc(C,A), savings(A)~=X), Y), linear([Y],[600,0.2],Mean).
creditScore(C) ~ gaussian(Mean,12.3) := client(C), \+mod(X,(hasAcc(C,A), freq(A)~=X),Z), max(X,(hasAcc(C,A), savings(A)~=X), Y), linear([Y],[500,0.8],Mean).
\end{lstlisting}
\end{example}

%\footnotetext{The \texttt{softmax} predicate implements the softmax function 
%$\texttt{Pj is} \  \frac{e^{w_{j_1} . V_1 + \cdots + w_{j_i} .V_i + w_{j_0}}}{\sum_{k=1}^{d}e^{w_{k_1} . V_1 + \cdots + w_{k_i} . V_i + w_{k_0}}}$. \nit{It together with the distribution in the head} signifies that a status $X$ \nit{of a loan} is sampled from a probability distribution $\left\{\frac{e^{0.1 \cdot Y - 0.3 \cdot Z - 2.4}}{N}\right\}^{1\left[X=appr\right]} \left\{\frac{e^{0.3 \cdot Y + 0.4 \cdot Z + 0.2}}{N}\right\}^{1\left[X=pend\right]}
%\left\{\frac{e^{0.8 \cdot Y + 1.9 \cdot Z - %2.9}}{N}\right\}^{1\left[X=decl\right]}$, where $N = e^{0.1 \cdot Y - 0.3 \cdot Z - 2.4} + e^{0.3 \cdot Y + 0.4 \cdot Z + 0.2} + e^{0.8 \cdot Y + 1.9 \cdot Z - 2.9}$.
%}

%\luc{ use the word attribute below once in a while}
At this point, it is worth taking time to study the above program in detail as several aspects of the probability distribution specified by the program can be directly read from it. First of all, the program specifies a probability distribution over $24$ random variables (cells) of the spreadsheet (Table \ref{tab:database}), 
%(fields) of the database, 
where $8$ of them belong to \texttt{client} table (age and credit score attributes of four clients), $8$ to \texttt{loan} table (loan amount and status attributes of four loans), and $8$ to \texttt{account} table (savings and frequency attributes of four accounts). When grounded, the set of clauses with the same head explicates random variables that directly influence the random variable defined in the head. For instance, the program explicates that the random variable \texttt{freq(a\_11)} directly influences the random variable \texttt{savings(a\_11)} since the distribution from which \texttt{savings(a\_11)} should be drawn depends on the state of \texttt{freq(a\_11)}. Similarly, the program explicates that random variables \texttt{freq(a\_11)}, \texttt{freq(a\_20)}, \texttt{savings(a\_11)} and \texttt{savings(a\_20)} directly influence the random variable \texttt{creditScore(ann)}, since the client \texttt{ann} has two accounts, namely \texttt{a\_11} and \texttt{a\_20}, and the credit score of \texttt{ann} depends on aggregate savings and aggregate frequency of these two accounts. The distributions in the head and the statistical models in the body of these grounded clauses quantify this direct causal influence. The program represents this knowledge about all random variables in a concise way.  

Unlike many graphical model-based representations such as PRMs \citep{getoor2001learning}, there is much local structure that is qualitatively represented by JMPs. To understand this point, let us reconsider clauses for credit score in Example \ref{example: examplePRT1}, the credit score of \texttt{ann} is independent of savings of all her accounts when \texttt{freq/1} (``frequency") of most of her accounts is \texttt{low} (a context). This is because in this context, the body of the last two clauses for the credit score can never be true and the first clause  specifies the distribution of \texttt{creditScore(ann)} without considering the states of savings of her accounts. To exploit these contextual independencies, the DC inference engine, which is based on probabilistic reasoning, finds proofs of the observation and query to determine the posterior probability of the query \citep{nitti2016probabilistic}. Note that PRMs construct ground Bayesian networks for inference, and it is well known that Bayesian networks can not qualitatively represent these independencies \citep{boutilier1996context}. \citep[][p. 239]{poole2008independent} provides a number of reasons for learning probabilistic logic programs.

\section{Learning Joint Model Programs}\label{section: learning}
%\luc{Luc's version}
The learning task consists of finding the hypothesis ${\cal H}$ that best explains the data ${\cal A}_{DB}$ w.r.t. the relational structure ${\cal R}_{DB}$ and the background knowledge ${\cal BK}$.  This setting is very much in line with traditional inductive logic programming \citep{lavrac1994inductive} and probabilistic inductive logic programming  \citep[PILP,][]{riguzzi2014history}.  It allows one to consider {\em background knowledge} about the entities and relations among the entities using a set of distributional clauses. As usual in inductive logic programming, we shall also use a declarative bias ${\cal L}$ to define which distributional clauses are allowed in hypotheses and a scoring function $score$ to evaluate the quality of candidate hypotheses. The declarative bias is quite standard, 
it is described in detail in \ref{appendix: declarative bias}.
%XXX. \luc{Move that part to the appendix}

\begin{figure}[t]
\centering
\includegraphics[width=1\textwidth]{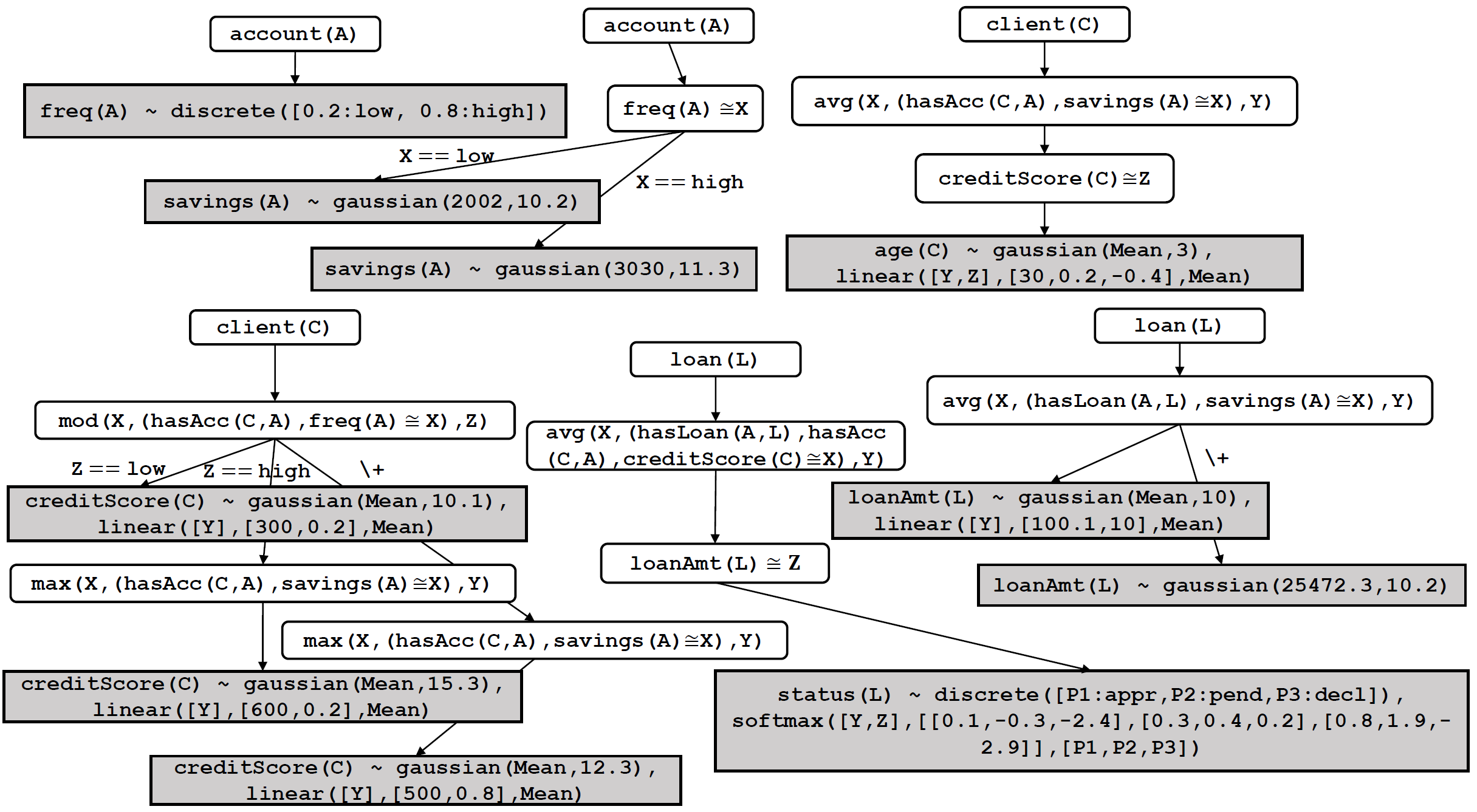}
\caption{A collection of DLTs corresponding to the JMP in Example \ref{example: examplePRT1}.}
\label{fig:ensembleOfTrees}
\end{figure}

%that can be expressed as the set of definite and/or distributional clauses $\mathcal{BK}$ as input. 

%The third component of the input is a {\em background knowledge}, that is, additional information about entities in the database and relations among the entities that the learning algorithm should take into consideration. 
%The set of facts and/or clauses $\mathcal{BK}$ describes background knowledge. 

%which can be any set of distributional clauses that express 
%satisfy the requirements on distributional clauses in JMPs (as described in Section \ref{section: JMP}). 
%The distributional clauses contained in the background knowledge must conform to the rand declarations, which are given as part of the language bias.
%The transformed database in the form of a distributional clause program along with the background theory and the declarative bias form the input of our learning algorithm. 

Rather than learning distributional clauses directly, we will learn {\em distributional logic trees} (DLTs), a kind of first-order decision trees  \citep{blockeel1998top} for distributional clauses. The reasons are 1) that decision trees are very effective from a machine learning perspective, and 2) that they automatically result in distributional clauses that are {\em mutually exclusive}, that is, they guarantee that the first validity requirement for DC is satisfied. This requirement states that only one distribution can be defined for each random variable in a possible world.

Formally, a DLT for an attribute, $a(T)$, is a rooted tree, where the root is an entity atom $e(T)$, each leaf is labeled by a probability distribution $\mathcal{D}_\phi$ and/or a statistical model $\mathcal{M}_\psi$, and each internal node is labeled with an atom $b_i$. Internal nodes $b_i$ can be of two types:
\begin{itemize}
    \item a binary atom of the form $a_j(T)\cong V$ that unifies the outcome of an attribute $a_j(T)$ with a variable $V$. %For instance, \texttt{freq(A) $\cong$ X}.
    \item an aggregation atom of the form $aggr(X,Q,V)$, as discussed in section \ref{sec:aggregation}, where $Q$ is of the form $(r(T,T_1), a_j(T_1) \cong X)$ in which $r$ is a link relation that relates entities of type $T$ to entities of type $T_1$ and $a_j(T_1)$ is an attribute.% For instance, \texttt{mod(X,(hasAcc(C,A),freq(A)$\cong$X),Z)}.
\end{itemize}
As common in decision trees, the nodes' children are defined based on the values that the node can take; here, this corresponds to the values that $V$ can take. There are two cases to consider:

%nit{Based on the type of the random variable defined by the internal nodes, the nodes can be of two types:}
%\luc{Pretty confusing -- you say above -- internal nodes can be of two types, and here you say it again, and use different "types"}
\begin{itemize}
\item  $V$ takes discrete values $\{v_1, ... , v_n\}$. Then there is one child for each value $v_i$.
\item  $V$ takes numeric values. Then its value is used to estimate the parameters of the distribution $\mathcal{D}_{\phi}$ and/or the statistical model $\mathcal{M}_{\psi}$ in the leaves. 
%These nodes can have two branches. Here also, the second branch is reserved for undefined. 
%Substitutions to the clause correspond to the path from the root to the node split at this node, and they choose the left branch of the node if the attribute is observed or the right branch if the attribute is not observed. All leaf nodes in the left branch of the node use the attribute to estimate the distribution. While leaf nodes in the right branch do not use it.
\end{itemize}
Furthermore, given that both the binary and the aggregation atom $b_i$ can fail, there is also an optional extra child that captures that $b_i$ fails and $V$ is undefined. This is reminiscent of logical decision trees, where every internal node contains a query, and there is both a success and a fail branch \citep{blockeel1998top}. Finally, the tree's leaf nodes contain the head of the distributional clause, which is of the form $h \sim \mathcal{D}_{\phi}$. The leaf node also includes the statistical model $\mathcal{M}_{\psi}$ present in the body of the distributional clause. 
%We now specify the distribution $\mathcal{D}_{\phi}$ and the model $\mathcal{H}_{\psi}$ that is present in our current implementation of \textit{LyRiC}. 

%In principle, any mathematical model that specifies a well-defined probabilistic model could be used. 
Depending on the type of the random variable defined by $h$, the distribution $\mathcal{D}_{\phi}$ and the model $\mathcal{M}_{\psi}$ can be one of the three types defined in Table \ref{table: dc with statistical models} in our current implementation of \dreaml. Examples of DLTs are shown in Figure \ref{fig:ensembleOfTrees}. It should be clear that if no continuous variable appears in the branch, then $\mathcal{M}_\psi$ is absent, and $\mathcal{D}_\phi$ is a Gaussian distribution or discrete distribution depending on the type of random variable defined by $h$. % Formally, 

It is straightforward to convert the DLT to a set of distributional clauses.  Basically, every path from the root to a leaf node in the DLT corresponds to a distributional clause of the form $h \sim \mathcal{D}_\phi \leftarrow b_1, ..., b_n, \mathcal{M}_\psi$.

%\luc{The reviewers asked for a formal definition of DLTs ... }

\begin{example}
The example of DLTs are shown in Figure \ref{fig:ensembleOfTrees}. There is one tree for each attribute, and together these DLTs make up the JMP of Example \ref{example: examplePRT1}.  Consider for instance the bottom-left DLT in the collection of DLTs shown in Figure \ref{fig:ensembleOfTrees}. The leftmost path from the root proceeding to the leaf node in the DLT corresponds to the following clause:
\begin{lstlisting}[frame=none]
creditScore(C) ~ gaussian(Mean,10.1) := client(C), mod(X,(hasAcc(C,A), freq(A)~=X),low), linear([Y],[300,0.2],Mean).
\end{lstlisting}
\end{example}

We can now summarize the learning task that is tackled by \dreaml as that of learning a DLT for a particular attribute. 
More formally, \\\\
\noindent
{\bf Given:} 
\begin{itemize}
    \item an attribute $a$
    \item training data consisting of,
        % n input DC program $\mathbb{P}_\mathcal{IN}$ consisting of,
    \begin{itemize}
        \item a set of facts $\mathcal{A}_\mathcal{DB} \cup \mathcal{R}_\mathcal{DB}$ representing a relational data
        %spreadsheet
        %database 
        $\mathcal{DB}$;
        \item a set of distributional clause (possibly empty) representing the background knowledge $\mathcal{BK}$;
    \end{itemize}
    \item a declarative bias ${\cal L}$ that defines the set of distributional clauses that are allowed in hypotheses;
    \item a scoring function 
    %$score(....)$ \luc{specify the arguments of score}
\end{itemize}
{\bf Find:} A distributional logic tree for $a$, which satisfies ${\cal L}$ and which scores best on the scoring function 

Once DLTs are learned for all attributes, they are converted to clauses that together with the set of facts $\mathcal{R}_{\mathcal{DB}}$ constitute the final learned JMP.

%with respect to $\mathcal{BK}$. \luc{You can also write : The DLT D for which $D = \arg\max_{D \in {\cal L}} score(...)$}
%best explains instances of the attribute $h$ and background knowledge, with respect to the scoring function.

%\luc{Say here that ADB is the training data, do not talk about input DC program, just about inputs to the DLT tree learner, and say again what the final 
%JMP will look like}

We now describe our approach \dreaml that learns JMPs. We do this in two different steps. We first present an algorithm to learn a DLT for a single attribute. Afterwards, we show how to learn a set of DLTs, that is, a JMP in an iterative EM-like manner, which is useful to deal with missing values.

\subsection{Learning a Distributional Logic Tree}\label{section: learning}

The distributional logic tree learner follows the standard decision tree learning algorithm sketched in Algorithm \ref{algorithm: induction of dlts}.
%It is sketched in Algorithm XX.

\begin{algorithm}[t]
\SetAlgoLined
\textbf{procedure} induce-DLT($\mathcal{T}:tree,~\mathcal{Q}:query,~\mathcal{E}: examples$,~${\cal V}: numeric ~ variables$)\\
%\Indp \textbf{Global} i) $a$: an attribute, ii) $\mathbb{P}_\mathcal{IN}$: an input DC program, iii) $\epsilon$: threshold \\
%\textbf{Input} i) $\mathcal{T}$: a DLT, ii) $\mathcal{Q}$: current body, iii) $\tau$: current score. \\
%\Indp In the start $\mathcal{T}, \mathcal{Q}$ are empty and $\tau$ is not a number. \\
%\Indm \textbf{Output} i) $\mathcal{T}$: a DLT \\
\Indp \textbf{if} $\mathcal{E} $ is not empty and sufficiently homogeneous {\bf then} \\
\Indp compute the best clause $a(T) \sim {\cal D}_\phi \leftarrow Q, {\cal M}_\psi$ using ${\cal V}$ according to $score$ \\
turn ${\cal T}$ into the leaf representing this clause\\
\Indm \textbf{else}\\
\Indp {\bf for} all queries $(Q,l(V)) \in \rho(Q)$ {\bf do}\\
\Indp compute $score((Q,l(V)),~{\cal E})$\\
\Indm {\bf end for}\\
let $(Q,l(V))$ be the best refinement with regard to the $score$\\
${\cal T}.test$ := $l(V)$\\
{\bf if } $V$ takes discrete values $\{v_1, ... ,v_n\}$ {\bf then}\\
\Indp {\bf for all} $v_i$\\
   \Indp  ${\cal E}_i$ := the set of examples in ${\cal E}$ for which $Q,l(V), V==v_i$ succeeds\\
   {\bf call} induce-DLT(${\cal T}.child(i), ~ {\cal E}_i,~(Q,l(V), V==v_i),~{\cal V})$  \\
   \Indm {\bf end for}\\
\Indm {\bf else} \ \%{\em $V$ takes continuous values} \\
      \Indp  ${\cal E}_v$ := the set of examples in ${\cal E}$ for which $Q,l(V)$  succeeds\\
{\bf call} induce-DLT(${\cal T}.child(success),~{\cal E}_v,~(Q,l(V)),~{\cal V} \cup \{V\})$  \\
\Indm  ${\cal E}_{fail}$ := the set of examples in ${\cal E}$ for which $Q,l(V)$ fails\\
%{\bf if} $E_{fail} \not= \emptyset$ {\bf then}\\
{\bf call} induce-DLT(${\cal T}.child(fail),~{\cal E}_{fail},~(Q,\backslash\mbox{+}~ l(V)),~{\cal V})$  \\
\caption{Induction of distributional logic trees}
\label{algorithm: induction of dlts}
\end{algorithm}

The induction process for the tree for a target attribute predicate $a(T)$ starts with the tree, and the query initialized to an entity predicate $e(T)$ of the same type as the attribute predicate, the full set of examples ${\cal E}$ and the empty set of variables ${\cal V}$. The algorithm recursively adds nodes in the tree. Before adding a node, it first tests whether the non-empty example set ${\cal E}$ is sufficiently homogeneous. If it is, it will compute the best statistical model ${\cal M}_\psi$ and the distribution ${\cal D}_\phi$ to be used in that leaf.  The set ${\cal E}$ is judged sufficiently homogeneous in a tree if none of the possible splitting or refinement operations increases the score by at least $\epsilon$. Furthermore, as there is no information in an empty set of examples,  the algorithm does not learn distributional clauses for branches of the tree that contain no examples.
 
In case the nodes should be further expanded, the standard recursive splitting procedure is followed, i.e., all possible tests $l$ to be put in the node are computed using a refinement operator $\rho$ and evaluated, the best refinement is selected and put in the node as a test, afterwards the children of the node are computed, and the procedure is called recursively. If the literal $l(V)$ produces discrete values, there is one branch per possible value; if it is continuous, there is one branch in which the value of the continuous variable $V$ will be remembered so that it can be used in the statistical model. The final branch is a fail branch corresponding to the case where the query $Q,l(V)$ fails. Such failing branches are also used in the logical decision tree learner TILDE \citep{blockeel1998top}. The process terminates when there are no attributes left to test on, or when examples at each leaf nodes are sufficiently homogeneous. 

%, it first tests whether the score of the clause corresponding to the root proceeding to the node increases when splitting the node, by at least a threshold $\epsilon$. If it does not, then the node is turned into a leaf node; otherwise, all possible refinements of the node are generated and scored using the scoring function in Equation \ref{eq: score}, which will be explained later in this section. The best refinement is selected and incorporated into the internal node. This internal node is split
%splitted 
%y adding new nodes to each of its branches. 
%and a new node is added to the tree for each branch of the internal node.\oknote{The last sentence is not clear. I think you want to say that you add child nodes to the node that is being split.} 
%This procedure is called recursively for the new nodes.
 
% That is, to learn the distribution ${\cal D}_\phi$ for an attribute $a(T)$,
% it will start from a tree with as root an entity atom $e(T)$ and then repeatedly decide
% whether 
%The key differences with a standard decision tree learning algorithm lies in the representation that is used.
%Essentially, every path in the distributional logic  tree corresponds to a query $Q$,  which in turns corresponds to the 
%%body of a distributional clause. Following standard inductive logic programming principles, we use a refinement operator $\rho$
%to generate the 

Several aspects of the algorithm still need to be explained in detail.

\paragraph{The refinement operator}
For generating refinements of the node, the algorithm employs a refinement operator \citep{dvzeroski2009relational}
%that works under $\theta$-subsumption \cite{de2008logical}. The operator
that specializes the body $\mathcal{Q}$ (the conjunction
of atoms in the path from the root to the node) by adding a literal $l$ to the body yielding $(\mathcal{Q}, l)$, 
where $l$ is either a binary atom of the form $a_j(T)$ or an aggregation atom as discussed in the beginning of this section. The operator ensures that only the refinements that are declarative bias conform are generated. %To ensure that more than one literal of the same attribute type does not appear in the body, once a literal corresponding to an attribute is added in the body, the refinement operator again does not add literals corresponding to the same attribute. This guarantees termination of the algorithm.
The details of the declarative bias are provided in \ref{appendix: declarative bias}.

%In the first part, we present an algorithm that learns JMPs from a spreadsheet
%%relational database 
%and background knowledge. 
%%This algorithm uses negated literals in the body of learned clauses to handle missing cells in the data. 
%In the second part, we present an iterative algorithm that learns the JMP by explicitly modeling the missing values, and that starts with the JMP learned in the first part. 

\paragraph{Estimating the parameters of the statistical model.}

The addition of the leaf node requires one to estimate parameters of the statistical model $\mathcal{M}_\psi$ and/or parameters of the distribution $\mathcal{D}_\phi$. Let us look at the following example to understand the estimation of the parameters.
\begin{example}
Suppose that the training data consists of the following set of facts and distributional clauses:
%input program $\mathbb{P}_\mathcal{IN}$ contains the following distributional clauses and facts:
\begin{lstlisting}[frame=none]
account(a_1).    account(a_2).
freq(a_1) ~ discrete([0.2:low,0.8:high]).    
freq(a_2) ~ val(low).
savings(a_1) ~ val(3000).    
savings(a_2) ~ val(4000).
deposit(A) ~ gaussian(30000, 100.1) := account(A), freq(A)~=low.
deposit(A) ~ gaussian(40000, 200.2) := account(A), freq(A)~=high.
\end{lstlisting}
Further, suppose that a path from the root to leaf node while inducing DLT for savings corresponds to the following clause, 
%Further suppose that we are interested in inducing DLT for savings and a path from the root to leaf node in the DLT corresponding to the following clause,
\begin{lstlisting}[mathescape=true,frame=none]
savings(A) ~ gaussian($\mu$,$\sigma$) := account(A), freq(A)~=low, deposit(A)~=X, linear([X],[$w_1$,$w_0$],$\mu$).  
\end{lstlisting}
where $\{w_0, w_1,\mu,\sigma \}$ are the parameters that we want to estimate. 

There are two substitutions of the variable \texttt{A}, i.e.,  $\theta_1 = \{\texttt{A/a\_1}\}$  and $\theta_2 = \{\texttt{A/a\_2}\}$, that are possible for the clause. The parameters of the clause can be approximately estimated from samples of the partial possible world obtained by proving the query $ \textrm{?- } h\theta_1, \mathcal{Q}\theta_1$ and the samples obtained by proving the query $\textrm{?- } h\theta_2, \mathcal{Q}\theta_2$. 
%Recall that the samples are drawn from the probability distribution $p(h\theta, \mathcal{Q}\theta)$ if we prove the query $ \textrm{?- } h\theta, \mathcal{Q}\theta$ in $\mathbb{P}_{\mathcal{IN}}$. 
%\oknote{Please see the comment in the chat...} \nit{NK: See my answer in the chat} 
Following Equation \ref{equation: probability weight}, the weight $w_{{\theta}_i}^{(j)}$ 
%$w^{(i)}$ 
of an $j^{th}$ sample obtained by proving a query $\textrm{?- } h\theta_i, \mathcal{Q}\theta_i$ is given by,
\begin{equation} \label{equation: proof weight}
%w^{(i)}
w_{\theta_i}^{(j)} = \frac{w_{q}^{(j)} w_{e}^{(j)}}{\sum_{j=1}^{N}w_{e}^{(j)}} 
\end{equation} 
%by proving the query $ \textrm{?- } h\theta_1, \mathcal{Q}\theta_1$ and the samples obtained by proving the query $\textrm{?- } h\theta_2, \mathcal{Q}\theta_2$. \oknote{Can we give more details here? Such as explicitly writing what the distribution is from which we are effectively sampling when proving the queries?} 
where $w_{q}^{(j)}$ is $1$ if the $j^{th}$ sample of the partial possible world entails the query;
%query $\textrm{?- } h\theta, \mathcal{Q}\theta$ is true in the $i^{th}$ sample; 
otherwise, it is $0$. Since the evidence set is empty, $w_e^{(j)}$ is always $1$ here.

%\nit{where $w_e$ is always $1$ since the evidence set is empty here.} 
Suppose, we obtained the following partial possible worlds, where each world is weighted by the weight obtained using Equation \ref{equation: proof weight}. 
\begin{lstlisting}[mathescape=true,frame=none]
[savings(a_1)~=3000,account(a_1),freq(a_1)~=low,deposit(a_1)~=30010.1],
${w}_{\theta_1}^{(1)}=0.5$.
[savings(a_1)~=3000,account(a_1),freq(a_1)~=high,deposit(a_1)~=40410.3],
${w}_{\theta_1}^{(2)}=0$.
[savings(a_2)~=4000,account(a_2),freq(a_2)~=low,deposit(a_2)~=30211.3],
${w}_{\theta_2}^{(1)}=0.5$.
[savings(a_2)~=4000,account(a_2),freq(a_2)~=low,deposit(a_2)~=30410.5],
${w}_{\theta_2}^{(2)}=0.5$.
\end{lstlisting}
\noindent
Thus, we have four data-points (i.e., partial possible worlds) to estimate parameters. The natural way for estimating the parameters is via log-likelihood maximization. However, in our case, each data-point is weighted. In such a case,  \cite{conniffe1987expected} argues that the estimation logically proceeds via expected log-likelihood maximization. So, to estimate the parameters, we maximize the expected log-likelihood of savings, that is given by the expression,
\begin{equation}\label{equation: expected log-likelihood example}
\begin{split}
   \ln(\mathcal{N}(3000 \mid 30010.1 w_1 + w_0, \sigma)) \times 0.5 + \ln(\mathcal{N}(3000 \mid 40410.3 w_1 + w_0, \sigma)) \times 0 + \\ \ln(\mathcal{N}(4000 \mid 30211.3 w_1 + w_0, \sigma)) \times 0.5 + \ln(\mathcal{N}(4000 \mid 30410.5 w_1 + w_0, \sigma)) \times 0.5
\end{split}
\end{equation}
%A good approximation of parameters can be obtained with only few samples in this simple example. 
It should be clear that the same approach can be used to estimate the parameters from any distributional clauses and/or facts present in the training data. 
%$\mathbb{P}_\mathcal{IN}$.

%Ideally infinite number of proofs are needed to get the exact estimate of parameters, since continuous random variables \texttt{deposit(a\_1)} and \texttt{deposit(a\_2)} can take infinite possible values. However, a good approximation of parameters can be obtained with only few proofs in this simple example. It should be clear that the same approach can be used to estimate the parameters for any distributional clauses and/or facts in $\mathbb{P}_\mathcal{IN}$.
\end{example}

%\oknote{Can we use the definition environment here?} 
Notice from the above example that substitutions of the clause are required to estimate the clause's parameters. We call such substitutions examples and define them formally, 
\begin{definition}({\em Examples at the leaf node}) Given the training data and a path from the root to a leaf node $L$ corresponding to a clause $h \sim \mathcal{D}_\phi \leftarrow \mathcal{Q}, \mathcal{M}_\psi$, we define the examples ${\cal E}$ at the leaf node $L$ to be the set of substitutions of the clause that ground all entity relations, link relations and attributes in the clause. 
\end{definition}
%It is worth emphasizing that $\Theta$ is not defined as the set of grounding substitutions of the clause since such grounding substitutions also ground the binary predicate $\cong$ that compares the outcome of a random variable with a value. 

Generalizing from Equation \ref{equation: expected log-likelihood example}, parameters of any distribution and/or of any statistical model at any leaf node can be estimated by maximizing the expected log-likelihood $E(\boldsymbol{\varphi})$, which is given by the following expression,
\begin{equation} \label{equation: expectation}
E(\boldsymbol{\varphi}) = \sum_{\theta_i \in \mathcal{E}} \sum_{j=1}^{N} \ln(p(h\theta_i \mid \boldsymbol{\varphi}, {\cal V}\theta_i^{(j)})){w}_{\theta_i}^{(j)}
\end{equation}
where $\boldsymbol{\varphi}$ is the set of parameters, ${\cal E}$ is the set of examples at the leaf node, ${\cal V}$ is the set of continuous variables in $\mathcal{Q}$, $N$ is the number of times the query $\textrm{?- } h\theta_i, \mathcal{Q}\theta_i$ is proved, ${w}_{\theta_i}^{(j)}$ is the weight of the $j^{th}$ sample, ${\cal V}\theta_i^{(j)}$ is $j^{th}$ sample of continuous random variables and $p(h\theta_i \mid \boldsymbol{\varphi}, {\cal V}\theta_i^{(j)})$ is the probability distribution of the random variable $h\theta_i$ given $\boldsymbol{\varphi}$ and ${\cal V}\theta_i^{(j)}$. For the three simpler statistical models that we considered, the expected log-likelihood is a convex function. \dreaml uses scikit-learn \citep{scikit-learn} to obtain the maximum likelihood estimate $\widehat{\boldsymbol{\varphi}}$ of the parameters. 

%\luc{what is the importance weight ? how does that relate to the weights in the formula's on the previous page ???? I do not get this .... }
%\luc{If this is the expectation -- how do you maximise it ??}
\paragraph{The Scoring Function}\label{sec:scoring}
%\luc{Should be written as $\sum_{C_I \in PDB}$}
%We consider the scoring function that is {\em decomposable}.
Clauses are scored using the Bayesian Information Criterion \citep[BIC,][]{schwarz1978estimating} for selecting the best among the set of candidate clauses. The score of a clause $h \sim \mathcal{D}_\phi \leftarrow \mathcal{Q}, \mathcal{M}_\psi$, which corresponds to a path from the root to a leaf, is given by, 
\begin{equation}\label{eq: score}
s(h \sim \mathcal{D}_\phi \leftarrow \mathcal{Q}, \mathcal{M}_\psi) = 2 E(\widehat{\boldsymbol{\varphi}}) - k\ln |{\cal E}|
\end{equation}
where $|{\cal E}|$ is the number of examples ${\cal E}$ at the leaf, $k$ is the number of parameters. The score avoids over-fitting and naturally takes care of the different number of examples at different leaves. To determine the score of the refinement $(\mathcal{Q},l(V))$ of the clause, where $V$ takes discrete values $\{v_1, \dots, v_n\}$, the score of $n+1$ clauses corresponding to $n+1$ branches are summed. That is, the score of the refinement is given by,
\begin{equation}\label{eq: score}
\begin{aligned}
score((\mathcal{Q},l(V)), \mathcal{E}_{V}) = s(h \sim \mathcal{D}_{\phi_1} \leftarrow \mathcal{Q}, l(V), V==v_1, \mathcal{M}_{\psi_1}) + \dots + \\
s(h \sim \mathcal{D}_{\phi_n} \leftarrow \mathcal{Q}, l(V), V==v_n, \mathcal{M}_{\psi_n}) + s(h \sim \mathcal{D}_{\phi_{fail}} \leftarrow \mathcal{Q}, \backslash\mbox{+}\ l(V), \mathcal{M}_{\psi_{fail}})
\end{aligned}
\end{equation}
where ${\cal E}_V$ is the number of substitutions to the clause $h \sim \mathcal{D}_{\phi_V} \leftarrow \mathcal{Q}, l(V), \mathcal{M}_{\psi_V}$. The score is computed in the similar manner when $V$ takes continuous value.

\paragraph{Learning Joint Model Programs}
To learn our final joint model program $\mathbb{P}_\mathcal{DB}$, we induce DLTs, in an order defined by the user in the declarative bias, separately for each attribute predicate. Recall that valid DC programs require the existence of a rank assignment $\prec$ over predicates of the program. The order declares the rank assignment over attributes. Each path from the root to the leaf node in each DLT corresponds to a clause in the program $\mathbb{P}_\mathcal{DB}$. 
%A collection of DLTs is obtained by separately inducing a DLT for each attribute predicate.
%%in the database. 
%Each path from the root node to a leaf node in each DLT corresponds to a clause in the learned JMP $\mathbb{P}_\mathcal{DB}$. 
This program defines the joint probability distribution
%spreadsheet, 
and probabilistic inference in this program can be used to compute a probability distribution over any set of cells given the observed value of any other set of cells.
%can now be used to infer the user's queries for the autocompletion task.

%In the experiment, we demonstrate that such programs achieves a state-of-the-art \citep{ravkic2015learning} performance during prediction. 
%The program obtained like this has the highest score; consequently, best explains the database and background knowledge.
%An example of the collection of DLTs is shown in Figure \ref{fig:ensembleOfTrees}. Learning the program $\mathbb{P}_\mathcal{DB}$ by inducing DLTs ensures that clauses in the program are mutually exclusive.
%\luc{I would not use the term ensemble here. In my intuitions ensembles are different predictive modesl for the same predicate / target variables, here it is for different targets. }

\subsection{Learning JMPs in the Presence of Missing Data} 
We explore two approaches in this paper:
\noindent
\subsubsection{Handling missing data using negated literals}\label{section: dc with negative literals}
One approach of learning probabilistic models from missing data that we have emphasized so far 
%One way of learning probabilistic models from data where some values may be missing 
is to treat missing values as a separate category and learn conditional distributions also for this category. By reserving one branch in the internal nodes for missing values (negation), DLTs do specify distributions for the target attribute ($a$) in the condition when values are missing. This branch corresponds to the {\em negated literal} in the distributional clause. 
\begin{example}
Consider DLT for \texttt{loanAmt} (``loan amount'') in the collection of DLTs shown in Figure \ref{fig:ensembleOfTrees}. The rightmost path from the root proceeding to the leaf node in the DLT corresponds to the clause with negated literal: 
\begin{lstlisting}[frame=none]
loanAmt(L) ~ gaussian(25472.3,10.2) := loan(L), \+avg(X, (hasLoan(A,L), savings(A)~=X), Y).
\end{lstlisting}
\end{example}
\noindent
The above clause specifies a distribution from which the loan amount is drawn if the loan has no account or the loan has accounts but the savings of these accounts are missing. 

There are other approaches of learning probabilistic models from missing data. The most common approach is Expectation-Maximization (EM). We discuss this approach next.
%EM in Section \ref{section: Stochastic EM}.

\subsubsection{Learning JMPs using the stochastic EM}\label{section: Stochastic EM}
%The program learned so far contains negated literals in the body of learned clauses to handle missing data. That is one approach to deal with missing data. Now we present another approach for learning programs in the presence of missing data. 
In this approach, we learn programs iteratively by explicitly modeling the missing data and start with the program learned so far. To realize this, we learn programs inside the stochastic EM algorithm \citep{diebolt1995stochastic}. In this setting, we assume that background knowledge is not present.

%In the second part, we present an iterative algorithm that learns the JMP by explicitly modeling the missing values, and that starts with the JMP learned in the first part. 

%In the previous section, we used negated literals in the body of learned clauses to deal with missing data.
%%fields in the database. 
%In this section, we discuss a more principled approach for learning programs in the presence of missing data.
%%fields. 
%This approach uses the stochastic EM \citep{diebolt1995stochastic} to deal with them.
%%the missing fields. 
%In this setting, we assume that background knowledge is not present.

Consider a training 
%input program $\mathbb{P}_\mathcal{IN}$ consisting of 
multi-relational tables
%a spreadsheet
%relational database
$\mathcal{DB}$ with missing cells
%fields 
$\mathbf{Z} = \{Z_1, \dots, Z_m\}$ and observed cells
%fields 
$\{X_1 \cong x_1, \dots, X_n \cong x_n\}$ (abbreviated as $\mathbf{X} \cong \mathbf{x}$), where $x_i$ is the value of the observed cell
%field
$X_i$. The iterative procedure starts by first learning a program $\mathbb{P}_\mathcal{DB}^0$ with negated literals from data with missing cells \textemdash{} using the same algorithm (Algorithm \ref{algorithm: induction of dlts}), subsequent programs are learned from data after filling missing cells with their sampled joint state. Formally, given the current learned program $\mathbb{P}_\mathcal{DB}^i$ specifying a probability distribution $p(\mathbf{X}, \mathbf{Z})$, the ($i+1$)-th EM step is conducted in two steps: 

\paragraph{E-step} A sample $\{Z_1 \cong z_1, \dots, Z_m \cong z_m\}$ (abbreviated as $\mathbf{Z} \cong \mathbf{z}$) of the missing cells
%fields 
$\mathbf{Z}$ is taken from the conditional probability distribution $p(\mathbf{Z} \mid \mathbf{X \cong x} )$. The missing cells
%fields
$\mathbf{Z}$ are filled in the tables
%spreadsheet
%database 
by asserting the facts $\{Z_1 \sim val(z_1), \dots, Z_m \sim val(z_m)\}$ (abbreviated as $\mathbf{Z} \sim val(\mathbf{z})$) in the training data.
%in $\mathbb{P}_\mathcal{IN}$. 

\paragraph{M-step} A new program ${\mathbb{P}_\mathcal{DB}^{i+1}}$ is learned from the training data
%input program $\mathbb{P}_{\mathcal{IN}}$ as discussed in section \ref{section: JMP with negations}
, and subsequently facts $\mathbf{Z} \sim val(\mathbf{z})$ are retracted from the training data.
%$\mathbb{P}_\mathcal{IN}$. 
However, in this case, the parameters of distribution and/or statistical models at the leaf node are estimated by maximizing the log-likelihood rather than maximizing the expected log-likelihood. This is because, in this case, the training data
%input program $\mathbb{P}_\mathcal{IN}$ 
does not consist of probabilistic facts or distributional clauses. Following equation \ref{equation: expectation}, the log-likelihood function $L(\boldsymbol{\varphi})$ is given by the following expression, 
\begin{equation}
L(\boldsymbol{\varphi}) = \sum_{\theta_i \in \mathcal{E}} \ln(p(h\theta_i \mid \boldsymbol{\varphi}, {\cal V}\theta_i^{(j)}))
\end{equation}

%The iterative procedure starts by first learning a program $\mathbb{P}_\mathcal{DB}^0$ with negated literals to deal with missing \nit{cells}.
%fields. 
%Subsequent programs learned after the first iteration do not contain negated literals since missing \nit{cells}
%fields 
%are filled in with their samples. 
The number of iterations decides the termination of the procedure. It is worth noting that we learn the structure as well as the parameters of the program $\mathbb{P}_{\mathcal{DB}}$, which is more challenging compared to learning only parameters of the model as in the case of standard stochastic EM. In the experiment, we demonstrate that the program learned at the end of stochastic EM procedure performs better compared to the learned program using the previous approach (Section \ref{section: dc with negative literals}).

%\oknote{Does the following make sense? (I am trying to explain the difference to the standard structural EM)}
The learning algorithm presented in this section is similar to the standard structural EM algorithm for learning Bayesian networks \citep{friedman1997learning}. The main difference, apart from having different target representations (DC vs. Bayesian networks), is that structural EM uses the standard EM \citep{dempster1977maximum} for structure learning. Our approach uses the stochastic EM for structure learning for the tractability reasons (hybrid probabilistic inference in large relational data
%databases 
is computationally very challenging).

\section{Experiments}\label{section: experiments}
This section empirically evaluates JMPs learned by \dreaml. We first describe the data sets that we used, and then explain the research questions that we address.

We used the same data sets as used in \cite{ravkic2015learning} to evaluate a hybrid relational model. Details of these data sets are as follows: 
\paragraph{Synthetic University Data Set} This data set contains information of $800$ \textit{students}, $125$ \textit{courses} and $125$ \textit{professors} with three attributes in the data set being continuous while the rest three attributes being discrete. For example, the attribute \textit{intelligence/1} represents the intelligence level of students in the range $[50.0, 180.0]$ and the attribute \textit{difficulty/1} represents the difficulty level of courses that takes three discrete values $\{easy, med, hard\}$. The data set also contains three relations: \textit{takes/2}, denoting which course is taken by a student; \textit{friend/2}, denoting whether two students are friends and \textit{teaches/2}, denoting which course is taught by a professor. 
%More details about the data set can be found in \cite{ravkic2015learning}.

\paragraph{Real-world PKDD'99 Financial Data Set} This data set is generated by processing the financial data set from the PKDD'99 Discovery Challenge. The data set is about services that a bank offers to its clients, such as loans, accounts, and credit cards. It contains information of four types of entities: $5,358$ \textit{clients}, $4,490$ \textit{accounts}, $680$ \textit{loans} and $77$ \textit{districts}. Ten attributes are of the continuous type, and three are of the discrete type. The data set contains four relations: \textit{hasAccount/2} that links clients to accounts; \textit{hasLoan/2} that links accounts to loans; \textit{clientDistrict/2} that links clients to districts; and finally \textit{clientLoan/2} that links clients to loans. 
%The original 
This data set is split into ten folds considering \textit{account} to be the central entity. All information about clients, loans, and districts related to one account appear in the same fold. 
%More information about the data set can be found in \cite{ravkic2015learning}. 
%More details about these data sets can be found in \cite{ravkic2015learning}.

%In addition to these benchmark data sets, we also performed experiments with one more data set whose results are reported in \ref{section: appendix}.

In addition to these benchmark data sets, we also performed experiments with one more data set:
\paragraph{Real-world NBA Data Set} 
This data set is about basketball matches from the National Basketball Association \citep{schulte2014aggregating}. It records information about matches played between two teams and actions performed by each player of those two teams. There are $30$ teams, $30$ games, $392$ players and $767$ actions. In total, there are $19$ attributes, and all of them are of integer type. We treated $18$ as continuous and $1$ attribute, i.e., \textit{resultofteam1/1} that takes two values $\{win, loss\}$ as discrete. This data set also contains relations, such as, \textit{team1id/2} that specifies the first team of matches, \textit{team2id/2} that specifies the second team of matches, \textit{teamid/3} that relates matches, players and teams. Considering the match to be a central entity, $90\%$ of the data set was used for training and the rest for testing.

Specifically, we address the following questions:

\begin{question}\label{section: experiment1}
\textbf{How does the performance of JMPs learned by \dreaml compare with the state-of-the-art hybrid relational models when trained on a fully observed data?}
\end{question}

We compared JMPs learned by \dreaml, in the case of fully observed data, with the model learned by the state-of-the-art algorithm \textit{Learner of Local Models - Hybrid (LLM-H)} introduced by \cite{ravkic2015learning}. The LLM-H algorithm learns a joint probabilistic relational model in the form of a hybrid relational dependency network (HRDN). %The algorithm requires the training data to be fully observed. In addition to LLM-H, we also compared the performance of JMP with individual DLTs learned for each attribute from the training data separately. 
This algorithm requires training data to be fully observed. To evaluate HRDNs, \citep{ravkic2015learning} followed the methodology of predicting an attribute of an instance in the testing data, using the rest of the testing data as observed. We followed the same methodology in this experimental setting. In addition to HRDNs, we also compared the performance of JMPs with individual DLTs learned for each attribute separately. Indeed on fully observed data, we could learn individual DLTs and use just one DLT to predict an attribute. However, then we could not deal with the autocompletion task, i.e., predicting any set of cells given any other set of cells. The current experimental setting, i.e., predicting a cell given all other cells, is simple compared to the autocompletion setting (our original problem). For clarity, we summarize the differences between these three models in Table \ref{table: differences}. 

\begin{table}[!ht]
\centering
 \begin{tabular}{|p{4cm}|p{4cm}|p{4cm}|} 
 \hline
 Individual DLTs & HRDNs & JMPs\\ [0.5ex] 
 \hline\hline
Individual models trained for individual attributes & Joint models specifying a joint probability distribution over all attributes & Joint models specifying a joint probability distribution over all attributes\\ \hline
Can make use of negated literals to deal with missing data & Can not deal with missing data & Can be trained using EM to deal with missing data and can also make use of negated literals\\ \hline
Can not be used for the autocompletion task that requires probabilistic inference & Can not be used for the autocompletion task\footnotemark & Can be used for the autocompletion task\\ \hline
\end{tabular}
\caption{Differences between individual DLTs, HRDNs, and JMPs.}
\label{table: differences}
\end{table}
\footnotetext{Although HRDNs are joint probabilistic models, inference in the presence of unobserved data, which is non-trivial, has not been studied (I. Ravkic, personal communication, February 2020). So it can not be used for the relational autocompletion task.}

Nonetheless, we performed this experiment as a sanity check to ensure that i) the individual DLTs that we learn are not worse than HRDNs and ii) the JMPs are not significantly worse than those DLTs. Even though we do not expect JMPs to be generally better since learning joint models has no advantage over learning individual models when training data is fully observed. Joint models can infer using both predictive and diagnostic information \citep{pearl2014probabilistic}, while individual models can only use predictive information. 
%joint models have no advantage over individual models when both training and testing data are fully observed. However, probabilistic inference in joint models can be used for autocompletion, while single models can not be used for such purposes.

We used the same evaluation metrics as used in \cite{ravkic2015learning} to evaluate the quality of predictions of JMPs. 

%\nit{We used the same data set, i) synthetic university data set and ii) real-world PKDD'99 financial data set as used in \cite{ravkic2015learning}.} These data sets are discussed in detail in Appendix \ref{section: appendix}. Also, we used the same evaluation metrics as used in \cite{ravkic2015learning} to evaluate the quality of predictions of JMP. 

\paragraph{Evaluation metric} To measure the predictive performance for discrete attributes, multi-class area under ROC curve ($\textrm{AUC}_{\textrm{total}}$) \citep{provost2000well} was used, whereas normalized root-mean-square error (NRMSE) was used for continuous attributes. The NRMSE of an attribute ranges from 0 to 1 and is calculated by dividing the RMSE by the range of the attribute. To measure the quality of the probability estimates, weighted pseudo-log-likelihood (WPLL) \citep{kok2005learning} was used, which corresponds to calculating pseudo-log-likelihood of instances of an attribute in the test data set and dividing it by the number of instances in the test data set.

\begin{table}[!ht]
\begin{center}
\scalebox{0.775}{
 \begin{tabular}{|c|c|c|c|c|c|c|} 
 \hline
 Evaluation & Predicate & HRDN & JMP & DLT\\ [0.5ex] 
 \hline
  \multirow{3}{*}{$\textrm{AUC}_{\textrm{total}}$} & gender/1 & 0.50 $\pm$ 0.01 & \textbf{0.52 $\pm$ 0.03} &  0.50 $\pm$ 0.03\\
  & freq/1 & 0.82 $\pm$ 0.01 & 0.77 $\pm$ 0.07 & \textbf{0.83 $\pm$ 0.04}\\
  & loanStatus/1 & 0.66 $\pm$ 0.04 & \textbf{0.82 $\pm$ 0.05} & 0.79 $\pm$ 0.04\\
 \hline
  \multirow{10}{*}{NRMSE} & clientAge/2 & 0.28 $\pm$ 0.02 & 0.24 $\pm$ 0.02 & \textbf{0.24 $\pm$ 0.01}\\ 
  & avgSalary/1 & \textbf{0.13 $\pm$ 0.02} & 0.18 $\pm$ 0.01 & 0.24 $\pm$ 0.00\\
  & ratUrbInhab/1 & 0.20 $\pm$ 0.00 & 0.25 $\pm$ 0.01 & \textbf{0.18 $\pm$ 0.00}\\
  & avgSumOfW/1 & \textbf{0.02 $\pm$ 0.00} & \textbf{0.02 $\pm$ 0.00} & 0.03 $\pm$ 0.01\\
  & avgSumOfCred/1 & \textbf{0.02 $\pm$ 0.00} & 0.02 $\pm$ 0.01 & 0.03 $\pm$ 0.01\\
  & stdOfW/1 & 0.05 $\pm$ 0.01 & 0.05 $\pm$ 0.01 & \textbf{0.05 $\pm$ 0.00}\\
  & stdOfCred/1 & 0.05 $\pm$ 0.01 & \textbf{0.04 $\pm$ 0.00} & \textbf{0.04 $\pm$ 0.00}\\
  & avgNrWith/1 & 0.15 $\pm$ 0.01 & \textbf{0.11 $\pm$ 0.00} & 0.11 $\pm$ 0.01\\
  & loanAmount/1 & 0.16 $\pm$ 0.02 & 0.12 $\pm$ 0.01 & \textbf{0.11 $\pm$ 0.01}\\
  & monthlyPayments/1 & 0.18 $\pm$ 0.02 & \textbf{0.14 $\pm$ 0.01} & 0.15 $\pm$ 0.02\\
 \hline
\end{tabular}}
\caption{The performance of JMP compared to HRDN and single trees for each attribute (DLT) on fully observed PKDD'99 financial data set. The best results (mean $\pm$ standard deviation) are in bold.}
\label{table:perfomanceOnFinancialDataset}
\end{center}
\end{table}

\begin{table}[!ht]
\begin{center}
 \begin{tabular}{|c|c|c|c|c|c|} 
 \hline
 Predicate & HRDN & JMP & DLT\\ [0.5ex] 
 \hline
nrhours/1 & -4.48 & -3.39 & \textbf{-3.20}\\  \hline
difficulty/1 & -0.02 & \textbf{-0.00} & -0.03\\ \hline
ability/1 & -5.34 & -3.83 & \textbf{-3.77}\\ \hline
intelligence/1 & -4.66 & -4.08 & \textbf{-3.37}\\ \hline
grade/2 & -1.45 & \textbf{-1.00} & \textbf{-1.00} \\ \hline
satisfaction/2 & -1.54 & \textbf{-1.05} & \textbf{-1.05}\\ \hline
Total WPLL & -17.49 & -13.35 & \textbf{-12.42}\\ \hline
\end{tabular}
\caption{WPLL for each attribute on fully observed university data set, consisting of 800 students, 125 courses, and 125 professors. The best results are in bold.}
\label{table:perfomanceOnUniversityDataset}
\end{center}
\end{table}

\begin{table}[!ht]
\begin{center}
\scalebox{1}{
 \begin{tabular}{|c|c|c|c|c|c|} 
 \hline
 Predicate & HRDN & JMP & DLT\\ [0.5ex] 
 \hline
plusminus/2		&	-5.38	&	-3.68	&	\textbf{-3.62} 	\\ \hline
defensiverebounds/2	&	-3.56	&	-2.14	&	\textbf{-2.12} 	\\ \hline
fieldgoalsmade/2		&	-1.66	&	\textbf{-0.58}	&	-1.03 	\\ \hline
assists/2			&	-3.10	&	-1.93	& 	\textbf{-1.91}	\\ \hline
blocksagainst/2		&	-1.36	&	-0.84	&	\textbf{-0.76}	\\ \hline
freethrowsmade/2		&	-1.52	&	-1.25	&	\textbf{-1.16}	\\ \hline
offensiverebounds/2	&	-2.27	&	\textbf{-1.36}	&	-1.41	\\ \hline
threepointattempts/2	&	\textbf{-0.00}		&	\textbf{-0.00}		&	\textbf{-0.00}		\\ \hline
threepointsmade/2		&	\textbf{-0.00}		&	\textbf{-0.00}		&	\textbf{-0.00}		\\ \hline
starter/2			&	-0.67	&	-0.70	&	\textbf{-0.36}	\\ \hline
turnovers/2		&	-2.45	&	-1.56	&	\textbf{-1.55}	\\ \hline
personalfouls/2		&	-2.44	&	-1.67	&	\textbf{-1.60}	\\ \hline
freethrowattempts/2	&	-1.66	&	\textbf{-0.98}	&	-0.99	\\ \hline
points/2			&	-2.87	&	\textbf{-1.84}	&	-1.90	\\ \hline
minutes/2			&	-10.91	&	\textbf{-7.21}		&	-7.21	\\ \hline
steals/2			&	-1.63	&	-1.03	&	\textbf{-1.03}	\\ \hline
fieldgoalattempts/2	&	-3.30	&	\textbf{-1.98}	&	-1.98	\\ \hline
blockedshots/2		&	-1.37	&	\textbf{-0.81}	&	\textbf{-0.81}	\\ \hline
resultofteam1/1		&	-2.05	&	\textbf{-0.00}		&	\textbf{-0.00}	\\ \hline
Total WPLL 		& 	-48.22 		&  	-29.56		& 	\textbf{-29.45}		\\ \hline
\end{tabular}}
\caption{WPLL for each attribute on the NBA data set. The best results are in bold.}
\label{table:perfomanceOnNBADataset}
\end{center}
\end{table}

In our experiment, we used the aggregation function \textit{average} for continuous attributes, and \textit{mode} and \textit{cardinality} for discrete attributes. An ordering chosen randomly among attributes was provided in the declarative bias. While training individual DLTs, ordering among attributes was not considered since those DLTs were not joint models but individual models for each attribute. We used the same data with the same settings as in \cite{ravkic2015learning} to compare the performance of our algorithm. Table \ref{table:perfomanceOnFinancialDataset} shows the comparison on financial data set using 10-fold cross-validation. %divided into ten folds. Nine folds were used for training and the remaining for testing. 
During testing, prediction of a test cell was the mode of the probability distribution of the cell obtained by conditioning over the rest of the test data. A Bayes-ball algorithm \citep{shachter2013bayes} that performs lazy grounding of the learned program was used to find the evidence that was relevant to the test cell. Table \ref{table:perfomanceOnUniversityDataset} shows the comparison on university data set divided into training and testing set. Numbers for HRDNs on these two data sets are taken directly from \cite{ravkic2015learning}. Table \ref{table:perfomanceOnNBADataset} shows the result on the additional data set, i.e., the NBA data set.

We observe that on several occasions, JMPs outperforms HRDNs, although both of these approaches use the same features to learn classification and regression models for attributes. This observation can be explained by the fact that LLM-H learns tabular conditional probability distributions (CPDs) while \dreaml learns tree-structured CPDs with much fewer parameters. \citep{chickering1997bayesian, friedman1998learning, breese1998empirical} observed that tree-structured CPDs are a more efficient way of automatically learning propositional probabilistic models from data. Unsurprisingly, we observe similar behavior for relational models as well. Apart from better performance, tree-structured CPDs make JMPs more interpretable. JMPs are human-readable programs while HRDNs are not. 
%Apart from better performance, JMP is more interpretable since it is a human-readable program, while LLH-H is a machine-readable program. 
As already discussed, we expect that single models for attributes, i.e., individual DLTs outperform both joint models, i.e., JMPs and HRDNs. It is worth reiterating that individual models can not be used for the autocompletion task, while joint models can be used. 
%In this experimental setting, both training and testing data were fully observed, so joint models had no advantage over single models. Probabilistic inference in joint models can be used to predict any set of cells given any other set of cells, while single models can not be used for such purposes.} 

The experiment suggests that JMPs learned by \dreaml can outperform the state-of-the-art algorithm for fully observed data. 

%of the same experiment as discussed for Question \ref{section: experiment1} when performed on this data set. 

%The result strengthens the conclusion made for Question \ref{section: experiment1}.

%Similar results were obtained on two other real-world data sets, 
%namely NBA and Hepatitis, 
%which are discussed in Appendix \ref{section: appendix}.

%Unsurprisingly, we observe that the performance of JMP is similar to individual DLTs since the training data were fully observed in this first experiment, consequently, single models for attributes, i.e., DLTs learned the same patterns in the data as JMP. Also, the test data were fully observed, so JMP had no advantage over individual DLTs. 
%Individual DLTs could not be used for prediction if some values were missing in the test data.  
%The performance of JMP is similar to LLM-H since LLM-H also uses the same features as \dreaml to learn classification or regression model for attributes; consequently, learning similar regularities from the data. However, on several occasions, JMP outperforms LLM-H.  The experiment suggests that JMP learned by \dreaml achieves comparable results as the state-of-the-art algorithm for fully observed data.

\begin{question}\label{section: experiment2} 
\textbf{Can \dreaml utilize background knowledge while learning distributional clauses? 
%learn distributional clauses utlizing bathe DLT for a single target attribute in the presence of background knowledge?
}
\end{question}

Background knowledge provides additional information about attributes that can be probabilistic when expressed as the set of distributional clauses. A learning algorithm that can utilize this information along with the training data can learn a better model. We performed this experiment to examine whether \dreaml can also learn a DLT for a single attribute (a set of clauses for an attribute) from the training data along with background knowledge expressed as a set of distributional clauses. This learning task is a more complex task than the previous task, where we learned individual DLTs from only training data, since this task involves probabilistic inference along with learning.

%Learning the DLT for a single target attribute from training data in the presence of background knowledge ($\mathcal{BK}$) is a more complex task compared to learning the DLT from only training data. $\mathcal{BK}$ provides additional information about attributes, so the learning task that involves probabilistic inference becomes complex. 
%In the logic programming literature, rule learners can induce clauses from training facts and background knowledge expressed as the set of clauses. One such example is ProbFOIL+ \citep{de2015inducing} for the Problog \citep{fierens2015inference}. With similar motivation, 
%We performed this experiment to examine whether \dreaml can also learn DLTs for a single attribute from the training data as well as the $\mathcal{BK}$ expressed as the set of distributional clauses.

\begin{figure}[!ht]
\centering
\hspace*{-1.2cm}
\includegraphics[width=1.19\textwidth]{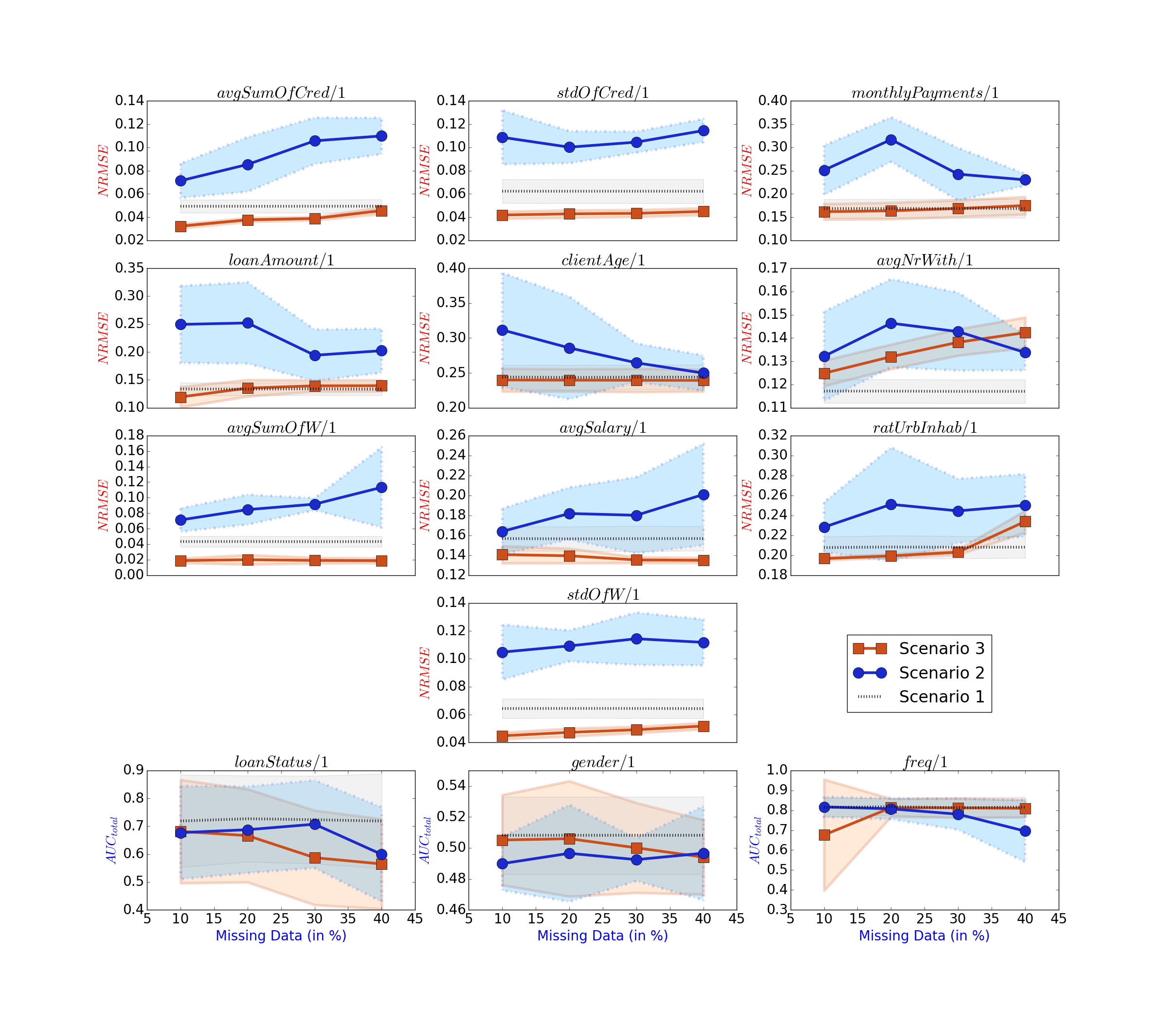}
\caption{Performance of models learned in the three scenarios (Question \ref{section: experiment2}) versus the percentage of removed cells. The bottom three figures show $\textrm{AUC}_{\textrm{total}}$ of discrete attributes, whereas,  the upper ten figures show NRMSE of continuous attributes. \underline{Less NRMSE} is better while \underline{more $\textrm{AUC}_{\textrm{total}}$} is better.}
\label{fig:results2b}
\end{figure}

We used the financial data set divided into ten folds. Two folds ($T$) were used for training the DLT for an attribute; one fold was used for testing that DLT; and seven folds were used for generating background knowledge $BK$, which was a set of distributional clauses for all attributes, i.e., a JMP. We considered three scenarios: 1) A DLT for an attribute was induced from the training set $T$; subsequently, the DLT was used to predict the attribute in the test fold. 2) A partial data set $T'$ was generated by removing $x\%$ of cells at random from the training set $T$. Subsequently, a DLT for the same attribute was induced from the partial set $T'$. Note that the DLT can be induced from partial data since we allow negated literals in the body of clauses. 
%In this case, some clauses in the DLT had negated literals in the body. 
%3) The missing fields in the partial set were filled with the mode of the probability distributions for the fields generated from the background knowledge; subsequently, a DT was induced for the same attribute. 
3) A DLT for the same attribute was induced from the partial set $T'$ as well as $BK$.

The predictive performance in the test set for the three scenarios, varying the percentage of removed cells, is shown in Figure \ref{fig:results2b}. Compared to the second scenario, much lower NRMSE is observed in the third scenario. On several occasions, DLTs learned in the third scenario, even outperform the same learned in the first scenario. Note that $BK$ is itself a probabilistic model learned from seven folds of data and is rich in knowledge. 

These results lead to the conclusion that \dreaml can learn distributional clauses from the training data utilizing additional probabilistic information from background knowledge.
%Similar results were obtained for other attributes that are discussed in Appendix \ref{section: appendix}.

%The predictive performance for $monthlyPayment/1$ in the test set in all three scenarios, varying the percentage of removed fields is shown in Figure \ref{fig:results2b}. Much lower NRMSE is observed in the third scenario. 
%Moreover, it performs better than scenario one when less data is removed from the training database. This is expected since the $\mathcal{BT}$ provides additional information about the removed fields and \textit{LyRiC} is able to use this information while learning DLT for the target attribute. 
%We can conclude that \dreaml can learn DLT from the training data \nit{utilizing} additional probabilistic information from $\mathcal{BK}$. 

%\subsection{\textbf{Can \dreaml learn the JMP from the relational data
%%database 
%when a large portion of the data
%%database 
%is missing?}}\label{experiment: EM}

\begin{question}\label{experiment: EM}
\textbf{Can \dreaml learn JMPs from relational data
%database 
when a large portion of the data
%database 
is missing?}
\end{question}

Probabilistic inference in a hybrid relational joint model is challenging. An even more challenging task, which requires numerous such inferences, is learning such models from partially observed relational data. We evaluated the performance of JMPs learned by \dreaml from such data. To the best of our knowledge, no system in the literature can learn such models from the partially observed relational data with continuous as well as discrete attributes. We used the financial data set and performed the following experiment to answer the question.

\begin{figure}[!ht]
\centering
\hspace*{-1.2cm}
\includegraphics[width=1.2\textwidth]{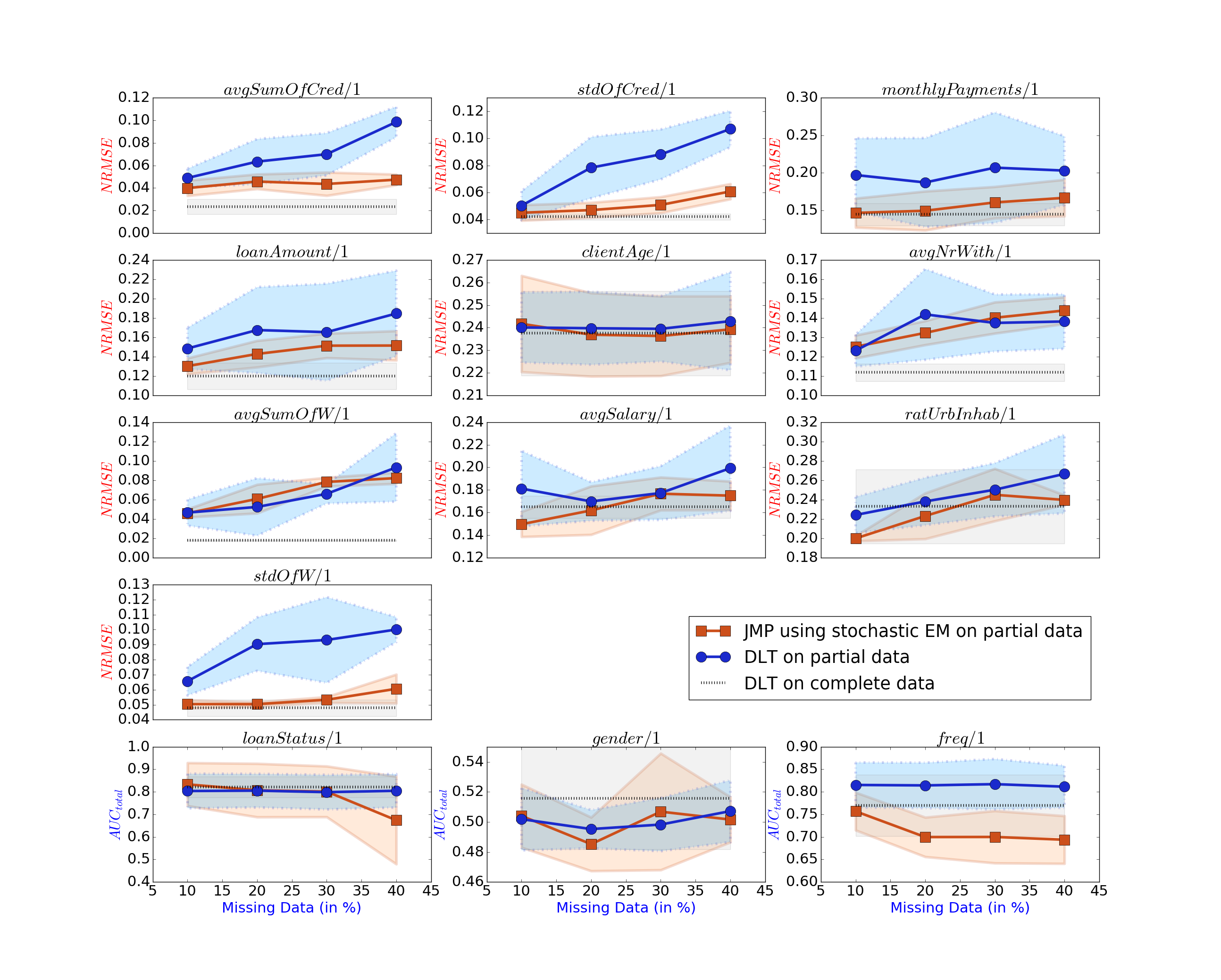}
\caption{Performance of the three models (Question \ref{experiment: EM}) on the financial data set versus the percentage of removed cells.  The bottom three figures show $\textrm{AUC}_{\textrm{total}}$ of discrete attributes, whereas,  the upper ten figures show NRMSE of continuous attributes. \underline{Less NRMSE} is better while \underline{more $\textrm{AUC}_{\textrm{total}}$} is better.}
\label{fig:results3c}
\end{figure}

\begin{figure}[!ht]
\centering
\includegraphics[width=0.7\textwidth]{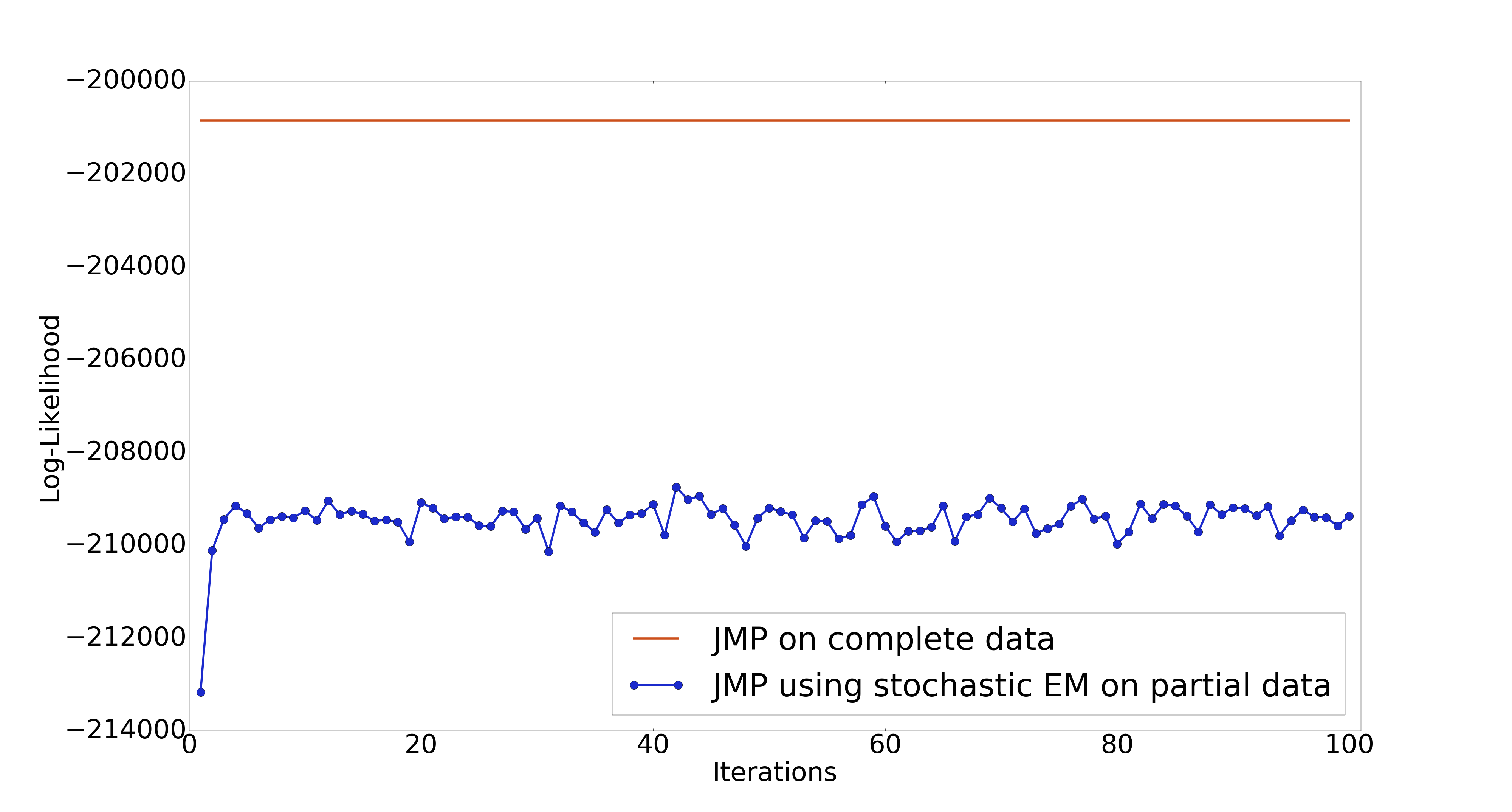}
\caption{The convergence of the stochastic EM on the financial data set}
\label{fig:results3e}
\end{figure}

We randomly removed some percentage of cells from the client, loan, account, and district tables of the financial data set to obtain a partial data set. Then we trained three models to predict attributes in the test data set.  The first model was a JMP obtained by performing stochastic EM on the partial data set. The second model was just an individual model, i.e., a DLT for each attribute trained on the partial data set. It is worth reiterating that the DLT can be learned even when some cells are missing since we allow negated literals in the body of distributional clauses. The last model was also an individual DLT for each attribute but was trained on the complete training data set. The performance of these models is shown in Figure \ref{fig:results3c}. Nine folds of the data set were used for training, and the rest for testing. The variance of NRMSE/$\textrm{AUC}_{\textrm{total}}$ is shown by shaded region when the experiment was repeated ten times on this data set. We observe that the JMP obtained using EM performs better, for most of the attributes, than individual DLTs trained on the partial data set. As expected, DLTs trained on the complete data perform best. The convergence of the stochastic EM after few iterations is shown in Figure \ref{fig:results3e}. To obtain this figure, the JMP was obtained from the financial data set with $10\%$ of cells removed using EM. This figure shows the data log-likelihood after each iteration of EM compared with the data log-likelihood when the JMP was obtained from the complete data.

The experimental environment was an Intel(R) Xeon(R) E5-2640 v3 2.60GHz CPU,  128GB RAM server running Ubuntu 18.04.4 LTS (64 bit). On the financial data set, \dreaml took approximately $226$ seconds to learn the JMP in each iteration of EM. The time required to sample a joint state of missing data from this program is shown in Table \ref{table:executionTimeOfEM}. 

\begin{figure}[!ht]
\hspace*{-1.2cm}
\includegraphics[width=1.2\textwidth]{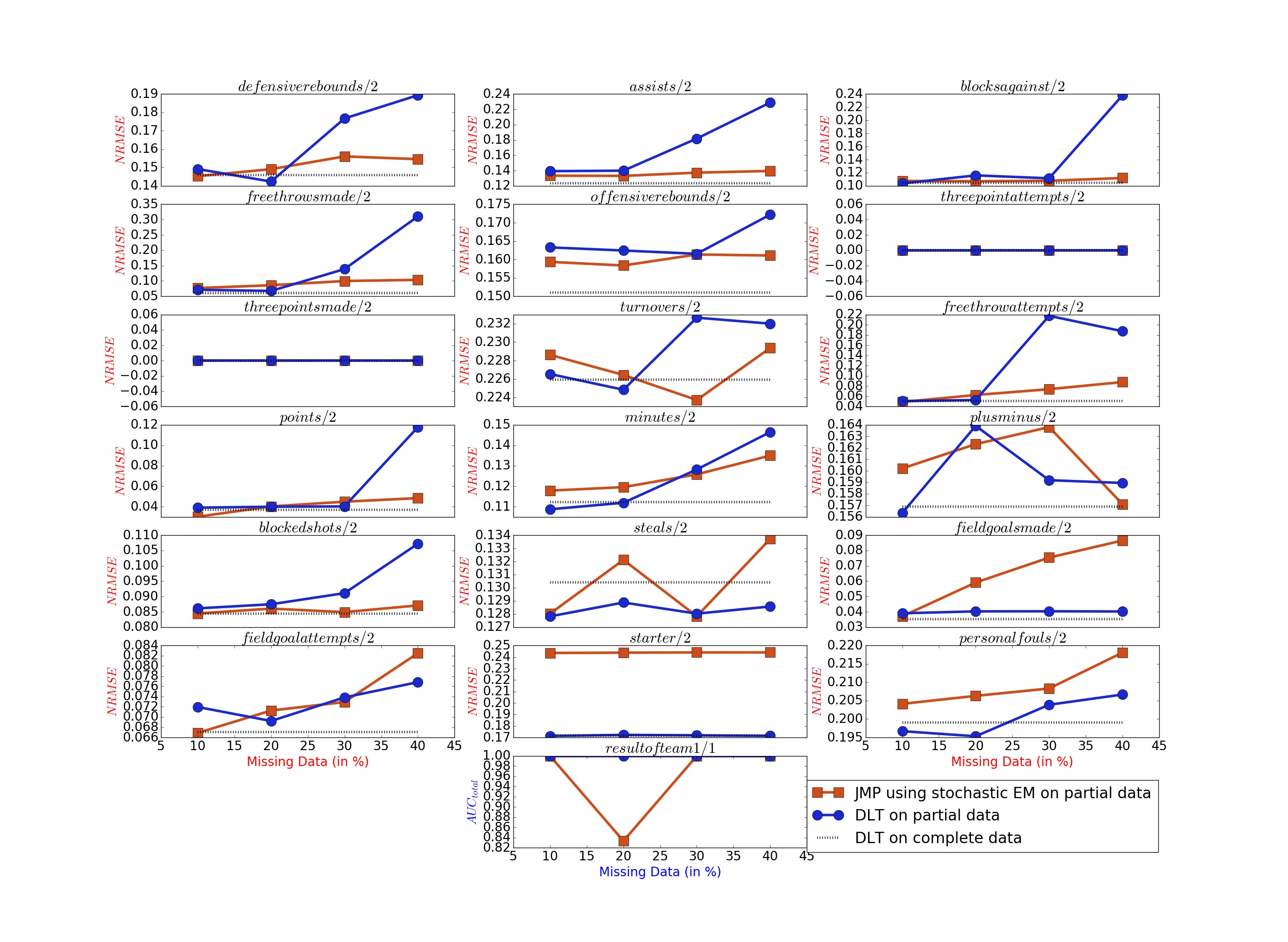}
\caption{Performance of the three models, discussed for Question \ref{experiment: EM}, on the NBA data set. The bottom figure show $\textrm{AUC}_{\textrm{total}}$ of the discrete attribute, whereas,  the upper eighteen figures show NRMSE of continuous attributes. \underline{Less NRMSE} is better while \underline{more $\textrm{AUC}_{\textrm{total}}$} is better.}
\label{fig:results3Nba}
\end{figure}

Results for the same experiment on the NBA data set is shown in
Figure \ref{fig:results3Nba}. We observe that when a large portion of data is missing, the JMP learned using stochastic EM performs better than individual DLTs. When $40\%$ of data is missing, the JMP performs better on $11$ attributes out of $19$ attributes. On $3$ attributes, the performance is the same. On $5$ attributes, individual DLTs perform better. 
%These attributes are at the bottom.

%To obtain the figure, first JMP was obtained from the complete PKDD financial database and using EM another JMP was obtained from the same database with $10\%$ of fields removed. This figure shows the database log-likelihood after each iteration of EM compared with the database log-likelihood obtained from the first JMP. 
%The comparison of the log-likelihood of the database after each iteration of EM with the log-likelihood of complete database is shown in this figure.

%To obtain the true log-likelihood \oknote{I would not call it ``true log-likelihood''.}, a JMP was obtained from the complete database. 
All these results demonstrate that \dreaml can learn JMPs even when a large portion of data is missing. 

\begin{table}[!ht]
\begin{center}
 \begin{tabular}{|c|c|c|} 
 \hline
 Percentage of missing cells & number of missing cells & Time in secs (approx.)\\ [0.5ex] 
 \hline\hline
10\%		&	3530	&  	131\\ \hline
20\%	&	7062	& 113	\\ \hline
30\%		&	10595	&	97	\\ \hline
40\%		&	14125	&	72	\\ \hline
\end{tabular}
\caption{The time taken to draw a joint state of missing data from the joint distribution.}
\label{table:executionTimeOfEM}
\end{center}
\end{table}

\section{Conclusions}\label{section: conclusion}

We presented \dreaml, a probabilistic logic programming based approach for tackling the problem of autocompletion in multi-relational tables.
%spreadsheets with multiple related tables. 
We first integrate distributional clauses with statistical models. Then these clauses are used to represent a hybrid relational model in the form of a DC program. Such a program is capable of defining a complex probability distribution over the entire related tables.
%spreadsheet. 
Probabilistic inference in this program allows predicting any set of cells given any other set of cells required by the autocompletion task. Since DC is expressive, we can map related tables
%entire spreadsheets 
to a set of facts in the DC language. In line with the approaches to (probabilistic) inductive logic programming, our approach learns such programs automatically from the set of facts and can make use of additional probabilistic background knowledge, if available. We demonstrated that such programs learned from fully observed relational data can outperform the state-of-the-art hybrid relational model. Another advantage of such programs over existing models is that such programs are interpretable. Although inference in hybrid relational models is hard, we demonstrated that the program learned by \dreaml performs well, even when a large portion of data is missing. \dreaml combines stochastic EM with structure learning to realize this.

\section*{Acknowledgements}
This work has received funding from the European Research Council (ERC) under the European Union’s Horizon 2020 research and innovation programme (grant agreement No [694980] SYNTH: Synthesising Inductive Data Models) and the Flemish Government under the “Onderzoeksprogramma Artificiële Intelligentie (AI) Vlaanderen” programme. OK's work has been supported by the OP VVV project {\it CZ.02.1.01/0.0/0.0/16\_019/0000765} ``Research Center for Informatics'', by the Czech Science Foundation project ``Generative Relational Models'' (20-19104Y) and a donation from X-Order Lab. Part of this work was done while OK was with KU Leuven and was supported by Research Foundation - Flanders (project G.0428.15).

\bibliographystyle{acmtrans}
\bibliography{main}

\appendix

\section{Declarative Bias}\label{appendix: declarative bias}
The use of {\em declarative bias}, which allows users to declaratively specify the search space of possible clauses to be explored while learning, is common in ILP systems such as PROGOL \citep{muggleton1995inverse}, TILDE \citep{blockeel1998top}, CLAUDIEN \citep{de1997clausal}, ALEPH \citep{srinivasan2001aleph}, etc. When the space is potentially huge, it plays an important role in restricting the search to finite and meaningful clauses. For our purposes, we adapt the bias declarations from \citep{de2008logical}. In \dreaml, the bias consists of four types of declarations, i.e., type,  mode, rand, and rank declarations. We describe them in turn with examples:

%As usual in inductive logic programming (ILP), we employ a syntactic bias to specify the type of variables in the clauses to be learnt. 
\paragraph{Types:} All functors are accompanied by {\em type declarations} of the form $type(func(t_{1}, \cdots, t_{n}))$, where $t_i$ denotes the type of the $i$-th argument, i.e., the domain of the variable. For instance, consider the type declarations in Figure \ref{fig:setting}. Since the first argument of \texttt{hasAcc/2} should be different type than the argument of \texttt{freq/1}, the clause
\begin{lstlisting}[frame=none]
age(C)~gaussian(30, 2.1) := mod(X,(hasAcc(C,A),freq(C)~=X),low).
\end{lstlisting}
is not type conform, but the following clause is:
\begin{lstlisting}[frame=none]
age(C)~gaussian(30, 2.1) := mod(X,(hasAcc(C,A),freq(A)~=X),low).
\end{lstlisting}
\paragraph{Modes:} We also employ modes, which is standard in ILP, for each attribute. Modes specify the form of literal $b_i$ in the body of the clause $h \sim \mathcal{D}_\phi \leftarrow b_1, \dots, b_n, \mathcal{M}_\psi$. A {\em mode declaration} is an expression of the form $mode(a_1, aggr, (r(m_{1},\dots, m_{j}), a_2(m_{k})))$, where $m_{i}$ are different modes associated with variables of functors, $aggr$ is the name of aggregation function, $r$ is the link relation, and $a_i$ are attributes. The expression specifies the candidate aggregation functions considered while learning clauses for the attribute $a_1$. If the link relation is absent, then the aggregation function is not needed, so the mode declaration reduces to the form $mode(a_1, none, a_2(m_{k})))$. The modes $m_i$ can be either {\em input} (denoted by ``$+$") or {\em output} (denoted by ``$-$").
%or {\em ground} (denoted by ``$c$"). 
The input mode specifies that at the time of calling the functor the corresponding argument must be instantiated, the output mode specifies that the argument will be instantiated after a successful call to the functor. Consider the mode declarations in Figure \ref{fig:setting}. The clause \begin{lstlisting}[frame=none]
age(C)~gaussian(30, 2.1) := mod(X,(cliLoan(C,L1),status(L2)~=X),appr).
\end{lstlisting} is not mode conform since the first argument of \texttt{cliLoan/2}, i.e., the variable \texttt{C} does not satisfy the output mode and the variable \texttt{L2} does not satisfy the input mode. The following clause, however, satisfies the mode:\begin{lstlisting}[frame=none]
age(C)~gaussian(30, 2.1):=mod(X,(cliLoan(C1,L1),status(L1)~=X),appr).
\end{lstlisting}
\paragraph{Rand Declarations:} They are used to define the type of random variables (i.e., discrete or continuous) and to specify the domain of discrete random variables.
%is defined by what we call {\em rand declarations}. 
\paragraph{Rank Declarations:} As we have already seen in Section \ref{section: distributional clauses}, the second validity condition of the DC program requires the existence of a rank assignment $\prec$ over predicates of the program. Hence, we introduce these declarations, 
%an additional declaration, which we call {\em rank declaration}, 
to specify the rank assignment over attributes. While learning distributional clauses for a single attribute, the rank declaration is not used, it is crucial while learning DC programs.
%\textit{LyRiC} requires that the user declares the bias.
%\luc{remarks about data wrangling distract from the main story -- delete}However, recent research in the area of automated data wrangling \cite{verbruggen2018automatically}, \cite{verbruggen2017towards}, \cite{contreras2018general} that aims to automatically learn declarative bias such as type declarations from the database looks promising. These methods could be used to automatically declare the bias for our learning algorithm.

\setttsize{\scriptsize}
\begin{figure}[t]
 \begin{adjustwidth}{-0.0cm}{}
\begin{tabular}{l | l}
\parbox{.5\textwidth}{
    \% Type declarations\\
    \texttt{type(client(c)).}\\
    \texttt{type(loan(l)).}\\
    \texttt{type(account(a)).}\\
    \texttt{type(hasAcc(c,a)).}\\
    \texttt{type(hasLoan(c,l)).}\\
    \texttt{type(age(c)).}\\
    \texttt{type(creditScore(c)).}\\
    \texttt{type(loanAmt(l)).}\\
    \texttt{type(status(l)).}\\
    \texttt{type(savings(a)).}\\
    \texttt{type(freq(a)).}\\
    \\
    \% Mode declaration \\
    \texttt{mode(age,none,creditScore(+)).}\\
    \texttt{mode(age,sum,(hasAcc(+,-),savings(+))).}\\
    \texttt{mode(age,avg,(hasAcc(+,-),savings(+))).}\\
    \texttt{mode(age,mod,(hasAcc(+,-),freq(+))).}\\
    \texttt{mode(age,max,(cliLoan(+,-),loanAmt(+))).}\\
    \texttt{mode(age,mod,(cliLoan(-,-),status(+))).}\\
    \texttt{mode(status,none,loanAmt(+)).}\\
    \texttt{mode(status,mod,(hasLoan(-,+),age(+))).}\\
    \vdots\\\\
    \% Rank declaration \\
    \texttt{rank([age,creditScore,loanAmt,}
}
&
\parbox{.5\textwidth}{
	\texttt{status,savings,freq]).}\\\\
	\% Random variable declaration \\
    \texttt{rand(age,continuous,[]).}\\
    \texttt{rand(creditScore,continuous,[]).}\\
    \texttt{rand(loanAmt,continuous,[]).}\\
	\texttt{rand(status,discrete,[appr,pend,decl]).}\\
    \texttt{rand(savings,continuous,[]).}\\
    \texttt{rand(freq,discrete,[low,high]).}\\
    \\
  	\% Transformed tables\\
  	\texttt{client(ann).}\\
  	\texttt{loan(l\_20).}\\
  	\texttt{account(a\_10).}\\
  	\texttt{age(ann) $\sim$ val(33).}\\
  	\texttt{creditScore(john) $\sim$ val(700).}\\
  	\texttt{savings(a\_10) $\sim$ val(3050).}\\
  	\texttt{freq(a\_10) $\sim$ val(high).}\\
  	\texttt{loanAmt(l\_20) $\sim$ val(20050).}\\
  	\texttt{hasAcc(ann,a\_11).}\\
  	\texttt{hasLoan(a\_11,l\_20).}\\
    \vdots\\
    \\
    \% Background knowledge\\
    \texttt{age(carl) $\sim$ gaussian(40,5.1).}\\
    \texttt{cliLoan(C,L)$\leftarrow$hasAcc(C,A),hasLoan(A,L).}
}
\end{tabular}
    \caption{An example of input to \dreaml, which consists of a 
    transformation of the spreadsheet
    %spreadsheet
    %relational database 
    in Table \ref{tab:database}, along with background knowledge and declarative bias.}
    \label{fig:setting}
 \end{adjustwidth}
\end{figure}
\setttsize{\small}

\begin{example} \label{example: input}
An example of the input to \dreaml is shown in Figure \ref{fig:setting}, where Table \ref{tab:database} is converted into facts, and background knowledge is expressed using distributional clauses.
%\luc{Do not talk about input DC programs any more ...} 
The first clause in the background knowledge shown in the bottom-right of the figure states that the age of \texttt{carl} follows a Gaussian distribution, and the second clause states that if a client has an account in the bank and the account is linked to a loan account, then the client also has a loan.  
%The declarative bias on the left is explained in the Appendix XXX.
%An example of the transformation of the relational database in Table \ref{tab:database} along with the declarative bias is shown in Figure \ref{fig:setting}. We could further assume to have additional domain knowledge, for instance, we know a prior that the probability distribution of the age of clients has normal distribution with mean $40$ and standard distribution $5.1$, and that if a client has an account in the bank and the account is linked to a loan account then the client also has a loan. This background knowledge can be then expressed as a set of clauses shown in the bottom-right of the figure. 
\end{example}

\label{lastpage}
\end{document}